\pdfoutput=1
\documentclass[10pt,twocolumn,letterpaper]{article}

\usepackage[top=0.9in,bottom=0.95in,left=0.75in,right=0.75in,columnsep=0.3in]{geometry}
\usepackage[T1]{fontenc}
\usepackage{times}

\usepackage{microtype}
\usepackage{graphicx}
\usepackage[font=small,labelfont=bf]{caption}
\usepackage{subcaption}
\usepackage{booktabs}
\usepackage{multirow}
\usepackage[numbers,sort&compress]{natbib}
\usepackage[colorlinks=true,allcolors=blue!55!black]{hyperref}
\usepackage{xurl}

\usepackage{amsmath}
\usepackage{amssymb}
\usepackage{mathtools}
\usepackage{xcolor}
\usepackage{colortbl}
\usepackage{stfloats} 
\usepackage[capitalize,noabbrev]{cleveref}
\usepackage{algorithm}
\usepackage{algorithmic}
\definecolor{tiereight}{HTML}{0072B2}
\definecolor{tierfour}{HTML}{009E73}

\newcommand{\ours}{ZenAttention}
\newcommand{\sagetwo}{SageAttention2++}
\newcommand{\sagethree}{SageAttention3}
\newcommand{\kvq}{LongLive~2.0 NVFP4 cache} 

\newcommand{\safefig}[2]{\IfFileExists{#2}{\includegraphics[width=#1]{#2}}{\fbox{\parbox[c][3cm][c]{#1}{\centering\small figure pending: \texttt{\detokenize{#2}}}}}}
\newcommand{\findone}{Finding~1}
\newcommand{\findtwo}{Finding~2}

\title{\textbf{ZenAttention: 4-bit Attention Directly on a Quantized KV Cache\\for Streaming Video Diffusion}}

\author{Yingwei Song$^{1,3}$ \qquad Tony Zhu$^{2}$ \qquad Kelly Peng$^{3}$ \qquad Yuan Gao \qquad Huanrui Yang$^{1,3}$\\[5pt]
{\normalsize $^{1}$University of Arizona \qquad $^{2}$Tsinghua University \qquad $^{3}$First Intuition}\\[3pt]
{\footnotesize\texttt{\{yingwei,huanruiyang\}@arizona.edu} \quad \texttt{me@tonyzoo.com} \quad \texttt{kelly@firstintuition.ai} \quad \texttt{williamgaoharvard@gmail.com}}}
\date{}

\begin{document}

\twocolumn[{%
  \renewcommand\twocolumn[1][]{#1}%
  \maketitle
  \vspace{-1.2em}
  \begin{center}
  \safefig{\textwidth}{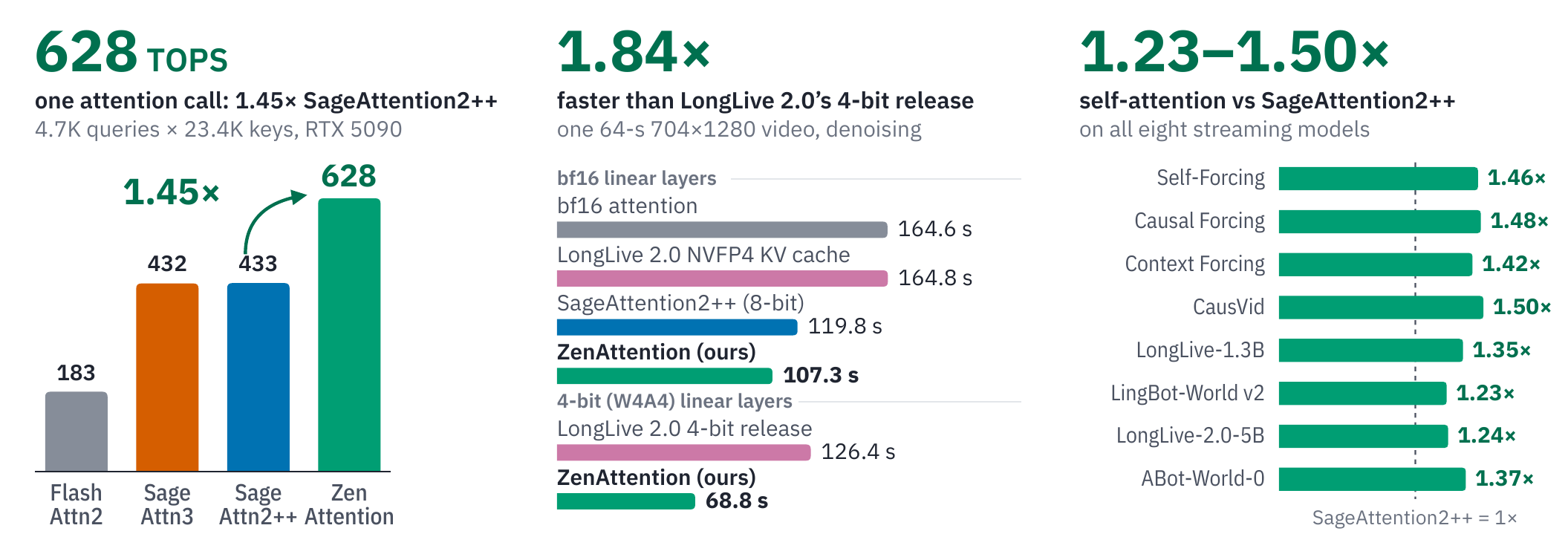}
  \captionof{figure}{\textbf{Speed of \ours{}} (RTX~5090). \emph{Left:} one attention call with 4,680 queries and 23,400 keys (Wan~2.1 1.3B, 12 cached frames). \emph{Middle:} denoising time of a 64-s $704\times1280$ video with LongLive-2.0-5B using its official inference pipeline; LongLive~2.0's NVFP4 cache is dequantized for bf16 attention, and its 4-bit release adds W4A4 linear layers. \emph{Right:} self-attention time per video, including cache writes, relative to \sagetwo{} (\cref{tab:speed}).}
  \label{fig:speed}
  \end{center}
  \vspace{0.6em}
}]

\begin{abstract}
In long-video generation, streaming video diffusion models spend up to 58\% of denoising time attending to a key--value (KV) cache of past frames at every step.
Keeping both the cache and the attention computation in 4-bit precision would save substantial memory and time, yet existing methods achieve only one of the two: FP4 attention requantizes a bf16 cache at every call and loses 4--28\% of high-frequency latent detail in streaming rollouts, whereas 4-bit KV caches dequantize back to bf16 before attention.
We trace the quality drop of 4-bit attention on a cache to two causes: FP4 attention error is systematically biased and fed back through the cache, amplifying single-chunk loss roughly $3\times$ on Self-Forcing; and the newest cached frame requires more precise attention scores than older history.
\ours{} addresses each cause with one component---closed-form bias corrections costing one fused multiply-add per score and an 8-bit tier for the newest frame---and computes attention directly on a write-once NVFP4 KV cache.

\ours{} is \emph{fast}: 628 effective TOPS on an RTX~5090 ($1.45\times$ \sagetwo{}), $2.6$--$3.5\times$ faster self-attention than bf16 ($1.23$--$1.50\times$ \sagetwo{}, faster than 8-bit and 4-bit SageAttention), and up to $1.89\times$ faster denoising than bf16 ($1.84\times$ over LongLive~2.0's 4-bit release);
\emph{general}: one kernel runs without retraining across eight streaming video and world models (1.3B--5B);
\emph{small}: KV-cache memory drops by two-thirds (33--35\% of bf16);
and \emph{faithful}: latent detail stays within 0.7\% of bf16 across all eight models (vs.\ 4--28\% lost by \sagethree{}), while VBench matches bf16 on Self-Forcing (946 prompts) and LongLive-2.0-5B (VBench-Long).

\noindent\textbf{Code:} {\small\url{https://github.com/wrencanfly/ZenAttention}}
\end{abstract}

\section{Introduction}
\label{sec:intro}

Streaming (autoregressive) video diffusion models generate long videos chunk by chunk \citep{yin2025causvid,huang2025selfforcing,yang2026longlive,chen2026longlive2}. At every denoising step, each chunk attends to the keys and values of earlier frames through a KV cache. With bf16 attention, this KV-cache self-attention accounts for 38--58\% of total denoising time (\cref{sec:speed,tab:speed}), making it the primary runtime bottleneck in long-video generation.

\textbf{Goal: both cache and attention at 4 bits.} Two properties of streaming video KV caches point directly to a 4-bit design. First, the cache is \emph{written once and read many times}: on Self-Forcing, each cached frame is read across up to 30 forward passes before leaving the sliding window. Second, every attention call processes thousands of queries (4,680 on Self-Forcing), making attention compute-bound on tensor-core throughput rather than memory-bandwidth-bound as in language-model decoding. On an RTX~5090, NVFP4 tensor cores execute an attention stage's matrix multiplications $1.98\times$ faster than 8-bit operands (\cref{sec:prelim-quant}), while the bf16 cache itself occupies 5.6--9.7\,GiB. Ideally, each finished chunk is quantized once to 4 bits and attended to directly in 4-bit arithmetic, saving both memory and compute.

\begin{figure*}[t]
  \centering
  \safefig{\textwidth}{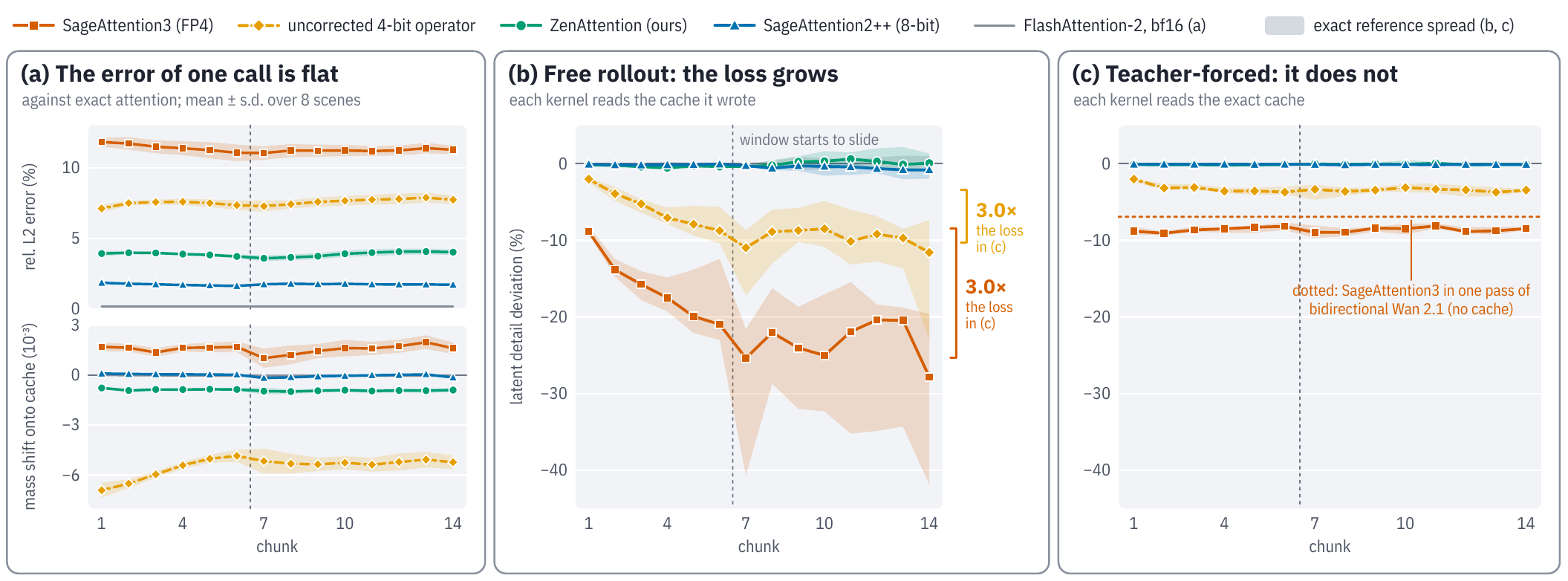}
  \caption{\textbf{FP4 errors compound through the cache} (Self-Forcing, 45 latent frames, 8 held-out scenes; \cref{sec:compound}). (a)~Relative $L_2$ error of one attention call and shift of attention mass onto cached keys, mean $\pm$ s.d.\ over scenes. (b, c)~Medians over scenes with interquartile bands; vertical dotted line: the window starts to slide. The brackets in (b) compare the loss in chunks 10--14 with the loss of the same kernel in (c), means over scenes; this gap is the compounded part.}
  \label{fig:compounding}
\end{figure*}

\textbf{Problem: existing methods realize only half of this design.} Current approaches split along the two halves (\cref{tab:options}). Existing FP4 attention, \sagethree{}~\citep{zhang2025sageattention3}, computes in 4 bits on a bf16 cache: requantizing the entire cached window at every call erases its speed advantage over 8-bit attention ($0.71$--$1.06\times$ per video), and in streaming rollouts it loses 4--28\% of high-frequency spatial detail in the latent space (\emph{latent detail}) across eight models (\cref{tab:models}). Conversely, 4-bit KV caches for video \citep{xi2026quantvideogen,lucas2026trq,zhao2026quantwm,chen2026longlive2} dequantize back to bf16 before attention: they save memory without speeding up compute, and LongLive~2.0's NVFP4 cache path is slower than plain bf16 attention (84.8 vs.\ 78.7\,s, \cref{app:longlive2}). Naively joining the two halves fails as well: uncorrected FP4 attention over a 4-bit cache still loses roughly 10\% of latent detail late in a Self-Forcing rollout (\cref{fig:compounding}).

\begin{table}[t]
\caption{\textbf{How existing methods treat the cached history.} ``Per call'': work performed on the full cached $K,V$ at each attention call. B/key: bytes per cached key and head ($K$, $V$, and all scales; $d=128$) on Self-Forcing.}
\label{tab:options}
\centering
\small
\setlength{\tabcolsep}{4pt}
\resizebox{\columnwidth}{!}{%
\begin{tabular}{@{}llllr@{}}
\toprule
Method & Cache & Per call & $QK^{\top}$\,/\,$\widetilde PV$ & B/key \\
\midrule
FlashAttention-2 & bf16 & --- & bf16\,/\,bf16 & 512 \\
\sagetwo{} & bf16 & quantize & INT8\,/\,E4M3 & 512 \\
LightX2V cache & 8-bit, once & --- & INT8\,/\,E4M3 & 258 \\
\sagethree{} & bf16 & quantize & NVFP4\,/\,NVFP4 & 512 \\
LongLive~2.0 cache & NVFP4, once & dequantize & bf16\,/\,bf16 & 144 \\
\rowcolor{tierfour!10}
\textbf{\ours{}} & NVFP4, once$^\ast$ & --- & NVFP4\,/\,NVFP4$^\ast$ & 177 \\
\bottomrule
\end{tabular}}
\\[2pt]{\footnotesize $^\ast$Includes the 8-bit record of the newest cached frame, computed in INT8\,/\,E4M3.}
\end{table}

\textbf{Why 4-bit attention on a cache fails.} We trace the degradation to two root causes (\cref{sec:analysis}).

\emph{\findone{}: FP4 attention error is biased, and the cache feeds it back.} Comparing free and teacher-forced rollouts on Self-Forcing separates single-call error from cache feedback: as long as every chunk reads an exact cache, the detail loss stays flat, but once a kernel's own keys and values enter the cache, the loss grows to roughly $3\times$ the single-chunk loss (\cref{fig:compounding}). The error fails to average out primarily because it is systematically biased by NVFP4 arithmetic: block-scaled rounding shrinks query and key vectors, zero-mean score noise inflates expected softmax weights via the exponential, and round-to-nearest on attention weights zeroes out small probabilities.

\emph{\findtwo{}: the newest cached frame requires more precise scores than older history.} In Wan~2.1 1.3B \citep{wan2025wan}, the backbone of six of our eight models, a single attention head inherited from the bidirectional base model concentrates nearly all its mass on the same spatial position of the previous frame, making it acutely sensitive to 4-bit score errors. On Self-Forcing, quantizing this single head to 4 bits reproduces the first-frame contrast drop of the entire FP4 operator, and restoring exact scores for this head alone eliminates it (\cref{fig:copyhead}). Across five streaming video models, keeping the newest cached frame at 8 bits consistently restores $0.07$--$0.47$ percentage points of first-frame contrast (\cref{sec:copyhead}).

\textbf{Our approach: one component per finding} (\cref{fig:workflow}). For \findone{}, \ours{} counters FP4 bias in closed form: queries and keys are split into exact means and FP4 residuals, scores are rescaled for shrinkage, half their estimated variance is subtracted, and attention weights are quantized with fixed dithered rounding. All corrections reuse statistics stored with the cache and cost one fused multiply-add per score. For \findtwo{}, the newest cached frame and current chunk form an 8-bit tier sharing a single online softmax with the 4-bit tier. Finished chunks are quantized once into NVFP4 \emph{records} and read directly by the kernel. Together, these two components cut the compounded detail loss on Self-Forcing from $6.10$ to $0.23$ percentage points inside the same emulation ($0.05$ with the compiled kernel, \cref{sec:compound}).

\textbf{Contributions.} Storing and computing on the cache directly in 4 bits unifies in one design what the methods of \cref{tab:options} achieve only separately---low-bit speed, low-bit cache memory, and bf16 quality. Specifically:
(i)~\emph{Diagnosis:} we separate single-call error from cache-feedback compounding via free versus teacher-forced rollouts and identify the Wan~2.1 previous-frame copy head behind newest-frame sensitivity (\cref{sec:analysis}).
(ii)~\emph{Kernel:} we build \ours{}, a compiled Blackwell CUDA kernel that combines closed-form bias corrections and an 8-bit newest-frame tier in one online softmax directly over write-once NVFP4 records (\cref{fig:workflow,sec:corrections}).
(iii)~\emph{Evaluation:} across eight streaming video and world models (1.3B--5B) without retraining (\cref{sec:setup}), \ours{} reaches 628 effective TOPS on an RTX~5090, speeds up self-attention by $2.6$--$3.5\times$ over bf16 and $1.23$--$1.50\times$ over \sagetwo{} (up to $1.89\times$ faster denoising over bf16; $1.84\times$ over LongLive~2.0's 4-bit release; \cref{fig:speed,sec:speed}), cuts KV-cache memory by two-thirds ($33$--$35\%$ of bf16; \cref{sec:e2e}), keeps latent detail within $0.7\%$ of bf16 across all eight models (\cref{tab:models}), and matches bf16 on VBench for Self-Forcing and LongLive-2.0-5B (\cref{tab:vbench,tab:vblong,sec:e2e}).

\textbf{Distinction from LLM KV-cache quantization.} Native low-bit KV-cache attention has been studied for LLM decoding \citep{du2026bitdecodingfp4,ni2026nvfp4mla,deng2026mpfa} and simulated in PyTorch for write-once NVFP4 caches \citep{gupta2026scalesearch}, while keeping recent tokens at higher precision echoes KIVI \citep{liu2024kivi}. Streaming video diffusion operates in a fundamentally different regime: multi-thousand-query attention is tensor-core-bound rather than bandwidth-bound, autoregressive rollout feeds quantization bias back into the cache across chunks, and \ours{} provides closed-form write-once bias corrections inside a compiled Blackwell FP4 kernel (\cref{sec:related}).

\section{Preliminary}
\label{sec:prelim}

\subsection{Streaming Video Diffusion with a KV Cache}
\label{sec:prelim-stream}

A streaming model generates latent frames in chunks of $F$ frames with $n$ tokens each. For the current chunk, let $Q,K_c,V_c\in\mathbb R^{N\times d}$ with $N=Fn$, and let $K_h,V_h$ hold the cached history; a window of $W$ frames includes the current chunk. Every self-attention layer computes $O=\mathrm{softmax}(Q[K_h;K_c]^{\top}/\sqrt d)\,[V_h;V_c]$. A chunk is denoised in $T$ steps; one further pass at noise level zero writes its $K_c,V_c$ into the cache, where they stay unchanged until they leave the window. On Self-Forcing \citep{huang2025selfforcing} ($F=3$, $n=1{,}560$, $W=21$, $T=4$), a call has $4{,}680$ queries and up to $32{,}760$ keys, and each cached chunk is read by up to $(T+1)(W/F-1)=30$ passes. Some models keep the first chunk permanently as a sink \citep{yang2026longlive}.

\subsection{FlashAttention}
\label{sec:prelim-fa}

FlashAttention \citep{dao2022flashattention,dao2024flashattention2} evaluates self-attention tile by tile with an \emph{online softmax}: each query tile tracks a running row maximum $m$, a weight sum $\ell$, and an output accumulator $O$, rescaling $\ell$ and $O$ by $e^{m-m'}$ whenever key tile $j$ raises the maximum to $m'=\max(m,\mathrm{rowmax}\,S_j)$ before accumulating $\widetilde P_j=e^{S_j-m'}$ and $\widetilde P_jV_j$, and finally normalizing $O/\ell$ (\cref{alg:zen}).

\subsection{Quantization}
\label{sec:prelim-quant}

Block quantization stores a tensor $X$ as low-bit codes paired with per-block scales; we write $\hat X=\mathcal Q_{\mathrm{fmt}}(X)$ for the resulting quantized tensor (whose numerical entries equal code $\times$ scale). Per-block INT8 uses scale $\max|X_{\mathrm{blk}}|/127$. NVFP4 \citep{alvarez2025nvfp4}, refining the microscaling formats of \citet{rouhani2023microscaling}, splits $X$ into 16-element blocks with one E4M3 scale $s=\max|X_{\mathrm{blk}}|/6$ per block and rounds $X/s$ to the E2M1 grid $\{0,\pm\tfrac12,\pm1,\pm\tfrac32,\pm2,\pm3,\pm4,\pm6\}$. Low-precision tensor cores multiply two quantized operands $(\hat A,\hat B)$ directly from their packed codes and scales without materializing bf16. On the RTX~5090, pure tensor-core loops of an attention stage ($QK^{\top}$ and $\widetilde PV$ on a $64\times64$ tile, $d=128$) run $1.98\times$ faster with NVFP4 operands than with the INT8 and FP16-accumulated E4M3 operands of \sagetwo{} (\cref{app:kernel}).

\textbf{Low-bit attention.} \sagetwo{} \citep{zhang2025sageattention2,zhang2025sageattention2pp} quantizes $Q,K$ to INT8 after subtracting the mean key (\emph{smoothing $K$}), and $\widetilde P,V$ to FP8. \sagethree{} \citep{zhang2025sageattention3} quantizes all four operands to NVFP4. Both keep the cache in bf16 and quantize it again at every call.

\textbf{Low-bit caches.} Two designs quantize the cache once when it is written. The persistent 8-bit cache of LightX2V \citep{lightx2v2025lightx2v} is read directly by an 8-bit kernel. The NVFP4 cache of LongLive~2.0 \citep{chen2026longlive2} is dequantized to bf16 before attention. \Cref{tab:options} lists what each method stores and what it does with the cached history at every call; none of them both stores the cache at 4 bits and multiplies it at 4 bits.

\section{Why 4-bit Attention on a Cache Fails}
\label{sec:analysis}

This section asks why FP4 attention loses detail in streaming generation and establishes two findings. \findone{}: the cache feeds per-call error back so the loss compounds (\cref{sec:compound}) because FP4 attention error is biased (\cref{sec:sources}). \findtwo{}: the newest cached frame needs more precise scores than the rest of the history (\cref{sec:copyhead}). \Cref{sec:method} answers each finding with one component.

\textbf{Metric.} A streaming rollout is chaotic: $0.17\%$ relative noise on every attention output sends it onto a different trajectory within one or two chunks, so a per-pixel comparison with a reference run measures divergence rather than quality. We therefore measure the latent \emph{detail} of each chunk $x$, $\|x-\mathrm{box}_{3\times3}(x)\|$, its high-frequency norm above a $3\times3$ spatial mean. Detail is given in percent relative to a reference run, and differences between methods in percentage points (\emph{points}). A \emph{scene} is a prompt with a seed. Detail detects blurring and washing out on any trajectory, and a deviation in either direction is an error: a loss is blur, a gain is oversharpening. Comparing an unperturbed reference run against a second run with $0.17\%$ independent relative noise on every attention output gives the noise floor (exact fp32 in this section, bf16 in \cref{sec:exp}; \cref{app:protocol}). In every run the first chunk goes through the reference kernel so all runs start alike (\cref{app:protocol}).

\subsection{\findone{}: The Loss Compounds Through the Cache}
\label{sec:compound}

\textbf{Setup.} An attention kernel can lose detail in two ways: through the error of the current call, and through the errors of earlier calls written into the cache. To separate the two, we run each kernel twice on each of 8 held-out Self-Forcing scenes. In a \emph{free} rollout the kernel's own keys and values enter the cache, as in normal use. In a \emph{teacher-forced} rollout the kernel still denoises every chunk but always reads the cache of the exact rollout, so its loss is the error of one chunk given a correct history. We compare four operators: \sagethree{}, \sagetwo{} and \ours{} as compiled kernels, and an \emph{uncorrected} FP4 operator, which quantizes queries and cached keys, attention weights, and values to NVFP4 with no correction and no 8-bit tier (emulated in fp32; the current chunk stays exact).

\textbf{Observation.} \Cref{fig:compounding} shows three things.
(i)~\emph{The error of one call does not grow.} The per-call error of every kernel is flat from chunk 1 to chunk 14: about $11\%$ for \sagethree{}, $7$--$8\%$ for the uncorrected operator, $3.6$--$4.1\%$ for \ours{} and $1.7\%$ for \sagetwo{}, while both uncorrected FP4 operators shift attention mass systematically relative to exact attention (\cref{fig:compounding}a, bottom).
(ii)~\emph{With an exact cache, the loss does not grow either.} The teacher-forced loss of detail is flat; in chunks 10--14 it is $8.5\%$ for \sagethree{} and $3.4\%$ for the uncorrected operator.
(iii)~\emph{With the kernel's own cache, the loss grows.} In free rollouts the loss grows until the 21-frame window starts to slide and then stays at about three times the teacher-forced loss: $25.3\%$ for \sagethree{} and $10.3\%$ for the uncorrected operator in chunks 10--14.

The cache therefore amplifies an error that is already present in one call; a single forward pass of bidirectional Wan~2.1 with \sagethree{} loses $6.3$--$7.9\%$ of latent detail. Both the uncorrected operator (7 of 8 scenes) and \sagethree{} (6 of 8 scenes) show flat per-call error alongside growing free-rollout loss whose compounded part exceeds the noise floor.

We call the free loss minus the teacher-forced loss, averaged over chunks 5--14, the \emph{compounded part}. It is $15.44$ points for \sagethree{}, $6.10$ points for the uncorrected operator, and $0.05$ points for \ours{} (and $0.23$ points when toggling only the corrections and 8-bit tier inside the same fp32 emulation, lower in 8 of 8 scenes; \cref{app:controls}).

\subsection{The Error of FP4 Attention Is Biased}
\label{sec:sources}

The loss in \cref{fig:compounding} is directional: both uncorrected FP4 operators lose detail within a single chunk and across the cache. For a quantized vector $\hat x=\mathcal Q_{\mathrm{FP4}}(x)$, three effects of NVFP4 arithmetic give the error a consistent sign.

\textbf{(1) Shrinkage.} Block-scaled rounding to the coarse E2M1 grid sends the small elements of a block to zero and so tends to shorten a vector along itself, $\langle\hat x,x\rangle=\alpha_x\|x\|^2$ with slope $\alpha_x<1$. A score $\langle\hat q,\hat k\rangle$ is then pulled towards zero by about $\alpha_q\alpha_k$, which flattens attention.

\textbf{(2) Jensen bias of the scores.} A zero-mean score error $\varepsilon$ with variance $\sigma^2$ raises the expected unnormalized weight, $\mathbb E\,e^{s+\varepsilon}\approx e^{s}e^{\sigma^2/2}$ \citep{tuncer2026quantizedkeys}. Keys with noisier scores therefore gain attention mass, and with a mixed-precision cache this shifts mass systematically between tiers.

\textbf{(3) Dead zone of $\widetilde P$.} Rounding to nearest sends every weight below $\tfrac14$ code unit ($1/24$ of the block maximum) to zero, which sharpens attention; with the unrounded sum as denominator the output also shrinks \citep{yu2026mxattention,anthony2026zyfa}.

\textbf{Empirical results.} With plain NVFP4 and no correction, Self-Forcing loses $4.4\%$ of the detail and $3.7\%$ of the norm within five chunks (\cref{tab:ablation}). The score-side effects (1)--(2) dominate and blur the output: quantizing only $QK^{\top}$ loses $3.6$ points more detail than quantizing only $\widetilde PV$, in 8 of 8 scenes, whereas (3) acts in the opposite direction and oversharpens once scores are unbiased (\cref{tab:ablation}).

\textbf{Is it the sign or the size that compounds?} The corrections of \cref{sec:corrections} remove the signed part of the error and halve its size ($7.5\%$ to $3.9\%$ per call). To see what each contributes, we inject errors into the exact attention output that match the magnitude of the uncorrected operator's error while altering its sign structure (\cref{app:controls}): the operator's own error \emph{negated}, the same error with \emph{randomized signs}, and Gaussian noise; the last two have zero mean. Three results follow.
(i)~\emph{The loss of one chunk comes from the sign alone.} With an exact cache, the zero-mean errors change detail by at most $0.15\%$, whereas the uncorrected operator loses $3.5\%$ and the negated error raises detail by $4.6\%$.
(ii)~\emph{Through the cache, the drift follows the sign.} In free rollouts the negated error makes detail grow by $17.9\%$, whereas the uncorrected operator loses $10.3\%$.
(iii)~\emph{The size compounds as well.} A zero-mean error loses almost nothing in one chunk yet loses $3.2\%$ of detail in free rollouts; its compounded part is $0.4$ of the uncorrected operator's ($2.35$ and $2.48$ against $6.10$ points).
Both the sign and the size of the error therefore compound, with sign dominating: projecting the error onto each head's low-pass and high-pass output components---only $2.0\%$ per-call error---compounds to $11.7$ points, nearly five times a $7.6\%$ zero-mean error (\cref{app:controls}). The corrections act on both: they remove the sign and halve the size.

\subsection{\findtwo{}: The Newest Frame Needs Precise Scores}
\label{sec:copyhead}

\begin{figure}[t]
  \centering
  \safefig{\columnwidth}{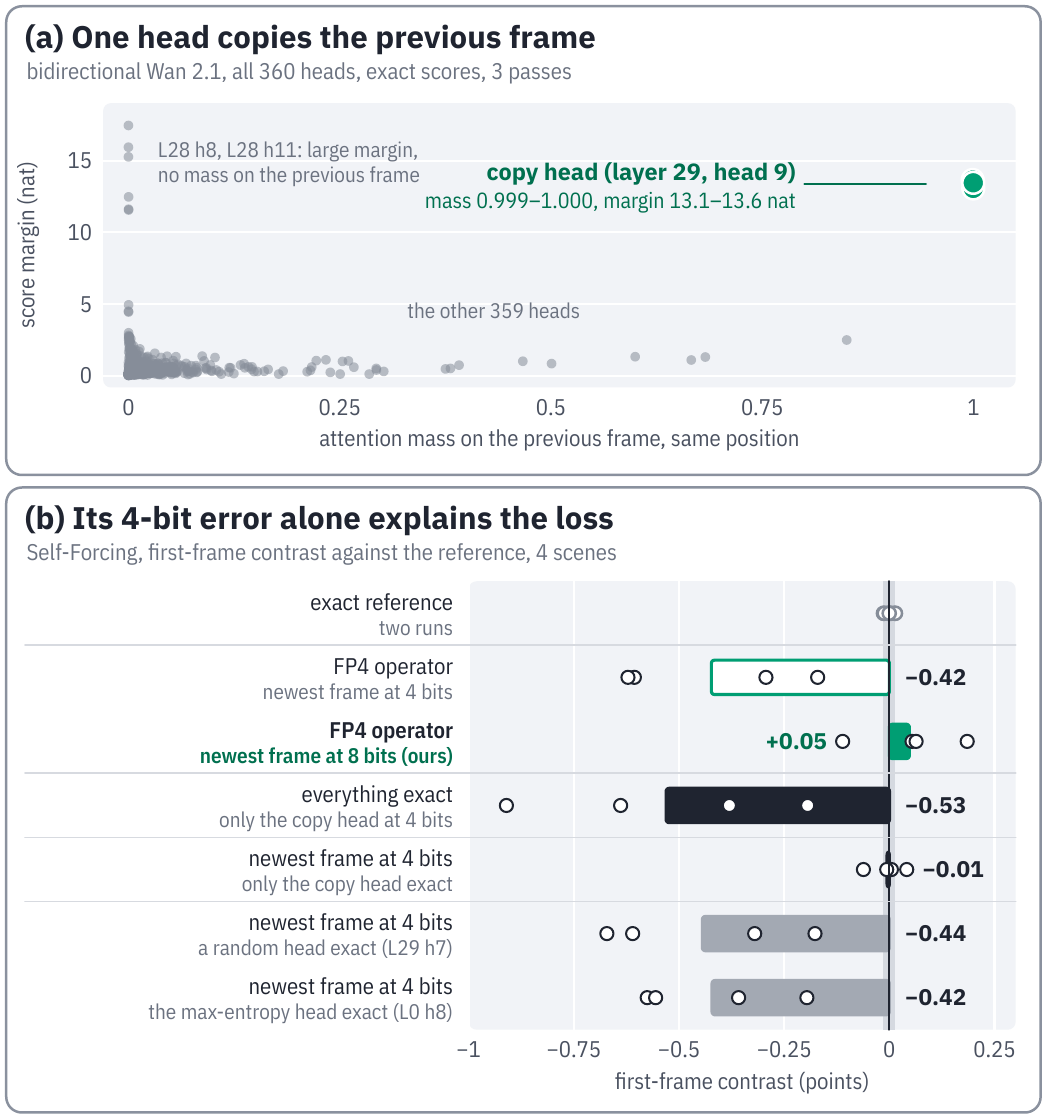}
  \caption{\textbf{The copy head} (Wan~2.1, layer 29, head 9). (a)~In the bidirectional model it is the only one of 360 heads that puts almost all its mass on the same position of the previous frame with a large score margin. (b)~First-frame contrast on Self-Forcing (4 scenes, FP4 tier emulated; bars: means, dots: scenes): this head's 4-bit error alone reproduces the loss of the FP4 operator without the 8-bit newest frame, and exact scores for this head alone remove it; the same for a random head or the highest-entropy head does not.}
  \label{fig:copyhead}
\end{figure}

The second failure is local: an FP4 operator loses detail in the \emph{first frame} of a chunk relative to the other frames. We measure it as \emph{first-frame contrast}, the detail of a chunk's first frame minus the mean detail of the remaining frames in the chunk.

\textbf{Observation.} In Wan~2.1 one head (layer 29, head 9) puts $0.999$ of its attention mass on the same position of the previous frame, with a score margin of about $13.5$ nats (\cref{fig:copyhead}a). It is present in bidirectional Wan~2.1 at several timesteps, so it is inherited from the bidirectional base model rather than introduced by causal distillation. For the first frame of every chunk, the previous frame is the newest cached frame. A head that copies from a single key is fragile at 4 bits: unlike isotropic Gaussian noise of the same size, its structured 4-bit score error can overturn the top key on a fraction of queries, acting like a mask on the few keys that carry the copy and causing an order of magnitude more damage (\cref{app:copyhead}).

\textbf{Empirical results.} On four emulated Self-Forcing scenes (\cref{fig:copyhead}b), the 4-bit error of this head alone, with everything else exact, loses $0.53$ points of first-frame contrast, against $0.42$ for the whole FP4 operator without an 8-bit newest frame. Exact scores for this head alone cut the loss to $0.01$. The same treatment of a random head (layer 29, head 7) or of the head with the highest mean entropy (layer 0, head 8) changes nothing. One head thus accounts for the loss, and more precise scores on the newest cached frame remove it: with that frame at 8 bits the contrast is $+0.05$.

\textbf{Across streaming models.} The copy head of Wan~2.1 is the sharpest instance of a general pattern: the newest cached frame carries most of the local temporal attention mass reading the cache, which is where 4-bit score errors cost detail. Across five streaming models, an 8-bit newest frame restores $0.07$--$0.47$ points of first-frame contrast (\cref{app:copyhead}). Likewise, across all eight streaming models of \cref{tab:models}, \sagethree{} loses $4$--$28\%$ of latent detail while \ours{} stays within $0.7\%$ of bf16, consistent with all three bias sources in \cref{sec:sources} arising directly from NVFP4 arithmetic.

\section{ZenAttention}
\label{sec:method}

\begin{figure*}[t]
\centering
\safefig{\textwidth}{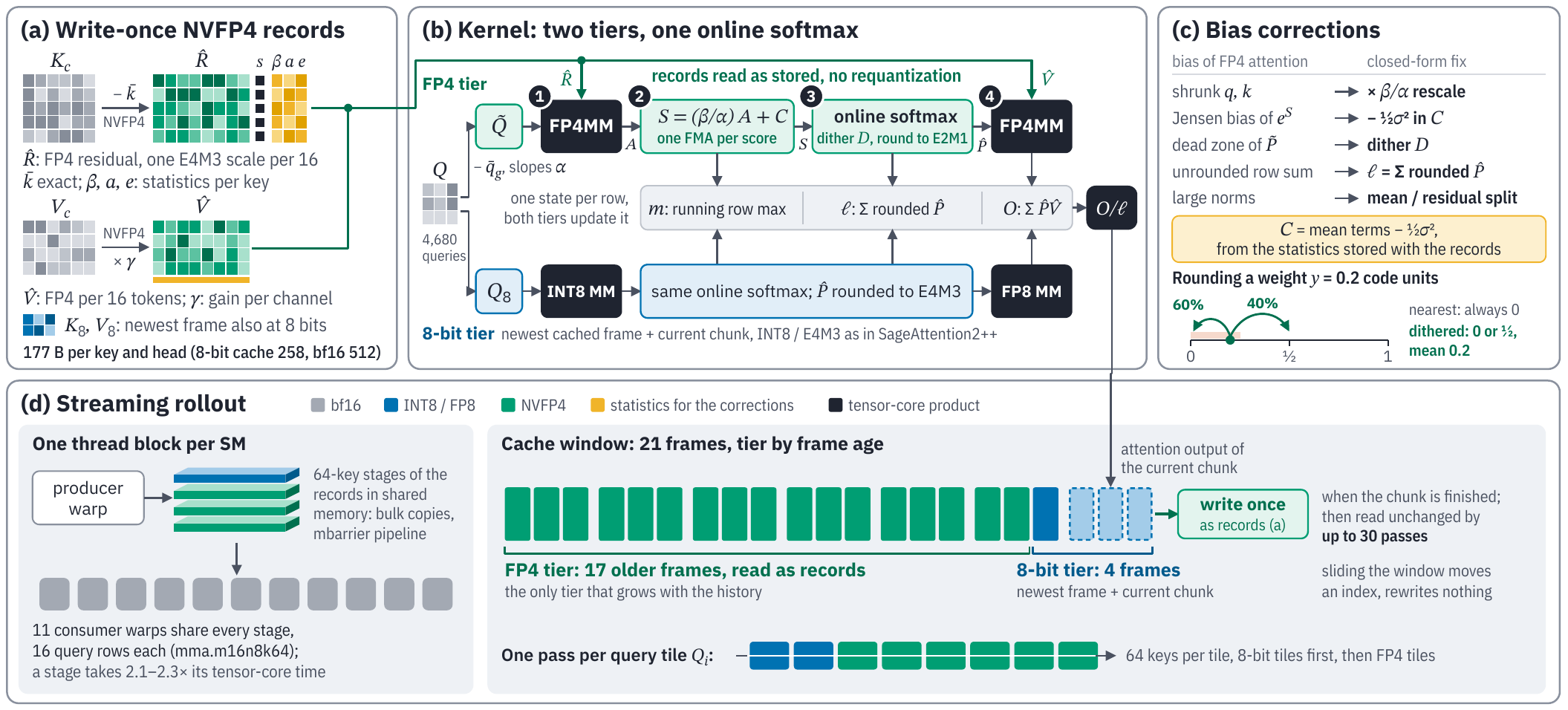}
\caption{\textbf{Overview of \ours{}.} (a)~A finished chunk is written once as NVFP4 records, with the statistics for the corrections (\cref{sec:records}). (b)~One call: an FP4 tier on the records and an 8-bit tier update one online softmax (\cref{sec:tiers}); steps 1--4 are lines 14, 15, 17 and 19 of \cref{alg:zen}. (c)~The corrections of \cref{eq:score} (\cref{sec:corrections}). (d)~The cache window of a rollout and the thread block of the kernel (\cref{sec:kernel}).}
\label{fig:workflow}
\end{figure*}

\ours{} keeps the design goal of \cref{sec:intro}, a cache that is written once at 4 bits and multiplied at 4 bits (\cref{sec:records}), and adds one component for each finding of \cref{sec:analysis}: closed-form bias corrections for \findone{} (\cref{sec:corrections}) and an 8-bit tier for the newest cached frame for \findtwo{} (\cref{sec:tiers}). \Cref{sec:kernel} describes the kernel, and \cref{fig:workflow} gives an overview.

\subsection{Write-once FP4 Cache Records}
\label{sec:records}

When chunk $c$ is finished, its keys and values are quantized once, from that chunk's own statistics, into records that the kernel reads directly. For keys, we keep the chunk mean $\bar k_c=\mathrm{mean}_j k_j$ in exact precision, quantize the residuals $r_j=k_j-\bar k_c$ to NVFP4 as $\hat r_j=\mathcal Q_{\mathrm{FP4}}(r_j)$ (row $j$ of $\hat R$), and store three scalars per key: the inverse shrinkage factor $\beta_j=\|r_j\|^2/\langle\hat r_j,r_j\rangle$, the quantized norm $a_j=\|\hat r_j\|^2$, and the residual error $e_j=\|\hat r_j-r_j\|^2$. Values are quantized in blocks of 16 tokens per channel and rescaled per channel by $\gamma=\sum v^2/\sum v\hat v_{\mathrm{raw}}$ into $\hat V$ to undo value shrinkage (analogous to per-channel norm compensation in \citet{banner2019posttraining}, here matching inner products along $v$). The newest frame of the chunk also gets an 8-bit record---INT8 keys $\hat K^8$ centered by that frame's mean $\bar k_f$ and E4M3 values $\hat V^8$---which the next chunk write replaces. Sliding the window updates an index table and rewrites nothing. On Self-Forcing the allocated cache, including the 8-bit newest frame, costs 177 bytes per key and head, against 258 for an 8-bit cache and 512 for bf16.

\subsection{Bias-corrected FP4 Attention (\findone{})}
\label{sec:corrections}

On the query side, each group $g$ of 128 queries is centered by its mean $\bar q_g$, and the residual $\tilde q_i=q_i-\bar q_g$ is quantized to NVFP4 as $\hat{\tilde q}_i=\mathcal Q_{\mathrm{FP4}}(\tilde q_i)$ with shrinkage slope $\alpha_i=\langle\hat{\tilde q}_i,\tilde q_i\rangle/\|\tilde q_i\|^2$. The FP4 tier scores key $j$ of chunk $c$ as
\begin{equation}
\begin{gathered}
s_{ij}=\tfrac{1}{\sqrt d}\Big(\tfrac{\beta_j}{\alpha_i}\langle\hat{\tilde q}_i,\hat r_j\rangle+\beta_j\langle\bar q_g,\hat r_j\rangle+\langle q_i,\bar k_c\rangle\Big)-\tfrac12\sigma^2_{ij},\\
\sigma^2_{ij}=\big(\|\hat{\tilde q}_i-\tilde q_i\|^2\,a_j+\|q_i\|^2\,e_j\big)/d^2 .
\end{gathered}
\label{eq:score}
\end{equation}
Only the first inner product runs on FP4 tensor cores; the remaining row and column terms are computed from stored vectors. \Cref{eq:score} and the rounding of $\widetilde P$ counter the three biases of \cref{sec:sources} with four techniques: the mean/residual split and scale correction counter shrinkage~(1), the variance correction counters Jensen bias~(2), and dithered rounding counters the dead zone~(3).

\textbf{Mean/residual split} (against shrinkage)\textbf{.} Quantizing only residuals and adding the mean products back reduces the operand norms that scale quantization error, extending the $K$ and $Q$ smoothing of \sagetwo{} and \sagethree{} \citep{zhang2025sageattention2,zhang2025sageattention3,lightx2v2025lightx2v} and the delta smoothing of \citet{li2026vcattention,li2026deltaquant}.

\textbf{Scale correction} (against shrinkage)\textbf{.} Multiplying by $\beta_j/\alpha_i$ makes the rescaled residuals $\hat{\tilde q}_i/\alpha_i-\tilde q_i$ and $\beta_j\hat r_j-r_j$ orthogonal to $\tilde q_i$ and $r_j$, matching the scale correction of DRIVE, EDEN and RaBitQ \citep{vargaftik2021drive,vargaftik2022eden,gao2024rabitq} without random rotations (\cref{sec:related}).

\textbf{Variance correction} (against the Jensen bias)\textbf{.} Under zero-mean isotropic quantization errors, the rescaled score before variance subtraction has approximate variance $\sigma^2_{ij}$, so subtracting $\tfrac12\sigma^2_{ij}$ cancels the first-order exponential inflation $\mathbb E\,e^{\tilde s_{ij}}\approx e^{s^{\star}_{ij}+\sigma^2_{ij}/2}$ (\cref{app:derivation}). Unlike prior corrections that assume exact queries and a parametric noise model \citep{tuncer2026quantizedkeys,dalberto2026kvcache}, $\sigma^2_{ij}$ is evaluated directly from the realized error norms of both operands stored in the records (\cref{sec:records}).

\textbf{Dithered rounding of $\widetilde P$} (against the dead zone)\textbf{.} On the uniform $\tfrac12$-step E2M1 sub-grid $[0,2)$, adding an offset $D_{ij}\in[-\tfrac14,\tfrac14)$ code units from a fixed $64\times32$ table (indexed by query, key block, and head) before nearest-neighbor rounding turns rounding into unbiased stochastic rounding \citep{gupta2015limited,croci2022stochastic}; each row is normalized by the sum of its \emph{rounded} weights \citep{yu2026mxattention,anthony2026zyfa}. Because NVFP4 rounding without random rotations is deterministic, the score and dither derivations hold only approximately (\cref{app:derivation}); \cref{tab:ablation,fig:compounding} validate each term directly in rollouts.

\subsection{An 8-bit Tier for the Newest Frame (\findtwo{})}
\label{sec:tiers}

\findtwo{} asks for more precise scores on the newest cached frame (\cref{sec:copyhead}). We therefore split the keys of a call into two tiers that share one online softmax (\cref{sec:prelim-fa}). The \emph{8-bit tier} holds the current chunk and the newest cached frame and uses INT8 $QK^{\top}$ and E4M3 $\widetilde PV$ as in \sagetwo{}. The \emph{FP4 tier} holds all older frames and uses the corrected scores of \cref{eq:score}. Frame age alone decides the tier, so no head or layer is treated specially. Both tiers update the same running maximum, rounded-weight sum and accumulator, so one normalization covers all keys (\cref{alg:zen}). Keeping the most recent tokens at higher precision follows \citet{liu2024kivi} and \citet{gupta2026scalesearch}.

\begin{algorithm}[t]
\caption{\ours{} for one head. [TC]: tensor-core product.}
\label{alg:zen}
\small
\begin{algorithmic}[1]
\STATE \textbf{Input:} current chunk $Q,K_c,V_c\in\mathbb R^{N\times d}$; newest-frame 8-bit record $(\hat K^{8},\hat V^{8},\bar k_f)$; older-frame FP4 records $\{\hat R,\bar k_c,\beta,a,e,\hat V\}$.
\STATE $\hat Q^{8}\!=\!\mathcal Q_{\mathrm{INT8}}(Q)$, $\hat K^{8}_c\!=\!\mathcal Q_{\mathrm{INT8}}(K_c)$, $\hat V^{8}_c\!=\!\mathcal Q_{\mathrm{E4M3}}(V_c)$
\STATE $\tilde Q=Q-\bar Q_{\mathrm{grp}}$; \ $\hat{\tilde Q}=\mathcal Q_{\mathrm{FP4}}(\tilde Q)$; \ slopes $\alpha$
\FOR{each query tile $i$ \textbf{in parallel}}
  \STATE $m_i=-\infty$, \ $\ell_i=0$, \ $O_i=0$
  \FOR{each key tile $j$ of the 8-bit tier}
    \STATE $S=\hat Q^{8}_i\hat K^{8\top}_j/\sqrt d$ [TC] \COMMENT{$+\,Q_i\bar k_f^{\top}/\sqrt d$ on newest frame}
    \STATE $m'=\max(m_i,\mathrm{rowmax}\,S)$; \ $\hat P=\mathcal Q_{\mathrm{E4M3}}(e^{S-m'})$
    \STATE $\ell_i\leftarrow e^{m_i-m'}\odot\ell_i+\mathrm{rowsum}\,\hat P$
    \STATE $O_i\leftarrow e^{m_i-m'}\odot O_i+\hat P\hat V^{8}_j$ [TC]; \ $m_i=m'$
  \ENDFOR
  \FOR{each key tile $j$ of the FP4 tier}
    \STATE $C=(Q_i\bar k_{c(j)}^{\top}+\beta_j\odot\bar q_g\hat R_j^{\top})/\sqrt d-\tfrac12\sigma^2_{ij}$ \COMMENT{row/column vectors}
    \STATE $A=\hat{\tilde Q}_i\hat R_j^{\top}$ [TC]
    \STATE $S=\tfrac{\beta_j}{\alpha_i}\odot A/\sqrt d+C$ \COMMENT{one FMA per score}
    \STATE $m'=\max(m_i,\mathrm{rowmax}\,S)$
    \STATE $\hat P=\mathcal Q_{\mathrm{FP4}}(e^{S-m'}+D_{ij})$ \COMMENT{dithered E2M1}
    \STATE $\ell_i\leftarrow e^{m_i-m'}\odot\ell_i+\mathrm{rowsum}\,\hat P$
    \STATE $O_i\leftarrow e^{m_i-m'}\odot O_i+\hat P\hat V_j$ [TC]
    \STATE $m_i=m'$
  \ENDFOR
  \STATE $O_i=O_i/\ell_i$
\ENDFOR
\STATE \textbf{return} $O=\{O_i\}$
\end{algorithmic}
\end{algorithm}

\subsection{Kernel Implementation}
\label{sec:kernel}

We implement \cref{alg:zen} in CUDA with inline PTX for compute capability 12.0, using the block-scaled FP4 instruction \texttt{mma.m16n8k64}: one launch prepares the queries (lines 2--3), one computes attention. The CUDA/PTX kernel, PyTorch reference, and drop-in model integrations are released at \url{https://github.com/wrencanfly/ZenAttention}.

\textbf{Corrections.} The query slope $1/\alpha_i$ and the softmax scale form one per-row exponent scale. The per-key scale $\beta_j$ and the additive terms $C$ (line 13) are applied with one fused multiply-add per score after the FP4 product, from per-row and per-key vectors stored with the records; the query statistics in $\sigma^2_{ij}$ are averaged over blocks of 64 queries. The dither adds one addition per score.

\textbf{Wide thread blocks.} One thread block per SM runs eleven consumer warps that share every 64-key stage, fed by a producer warp through an \texttt{mbarrier} pipeline (\cref{app:kernel}).

\textbf{Where the time goes.} Besides 33 tensor-core instructions per consumer warp and stage, the FP4 loop issues about 418 others for exponentials, maxima, quantizing, dithering and packing $\widetilde P$ (\cref{app:kernel}). A stage takes $2.3$--$2.5\,\mu$s, $2.1$--$2.3$ times its pure tensor-core time of $1.08\,\mu$s, so the gap to raw FP4 tensor-core throughput is dominated by per-score scalar work rather than memory traffic (alternative schedules change runtime by under $2\%$, \cref{app:kernel}). The 8-bit tier takes $23\%$ of the kernel time. The kernel matches a PyTorch reference of \cref{alg:zen} to $10^{-4}$ relative error.

\section{Experiments}
\label{sec:exp}

\begin{table}[t]
\caption{\textbf{VBench on Self-Forcing.} Flagged: cells of 21 (7 dimensions $\times$ whole, first, and last third) flagged against bf16 (\cref{app:protocol}). Max $|\Delta|$: largest whole-video difference from bf16 over the six continuous dimensions (VBench score $\times100$). Full scores: \cref{app:vbench}.}
\label{tab:vbench}
\centering
\small
\setlength{\tabcolsep}{4.5pt}
\begin{tabular}{lcccc}
\toprule
 & \multicolumn{2}{c}{11\,s, $n=946$} & \multicolumn{2}{c}{60\,s, $n=128$} \\
\cmidrule(lr){2-3}\cmidrule(l){4-5}
 & Flagged & Max $|\Delta|$ & Flagged & Max $|\Delta|$ \\
\midrule
\sagetwo{} & 0/21 & 0.02 & 0/21 & 0.42 \\
\sagethree{}$^\ast$ & 10/21 & 5.0 & 14/21 & 18.8 \\
\rowcolor{tierfour!10}
\textbf{\ours{}} & \textbf{0/21} & \textbf{0.03} & \textbf{0/21} & \textbf{0.15} \\
\bottomrule
\end{tabular}
\\[2pt]{\footnotesize $^\ast$Evaluated on 128 prompts at 11\,s and 32 prompts at 60\,s.}
\end{table}

\begin{table*}[t]
\caption{\textbf{Latent detail on eight models}: change against the bf16 baseline over the whole video (\%), mean over $n$ prompts, for \sagetwo{} (SA2++), \sagethree{} (SA3) and \ours{} (Ours). Ours $-$ SA2++ is paired by prompt; bold: the 95\% interval excludes zero. bf16 spread: half the range of a prompt's bf16 runs. Speed per model: \cref{fig:speed} and \cref{tab:speed}.}
\label{tab:models}
\centering
\small
\setlength{\tabcolsep}{6pt}
\begin{tabular}{lrrr>{\columncolor{tierfour!10}}rrcc}
\toprule
Model & $n$ & SA2++ & SA3$^\ast$ & \textbf{Ours} & Ours $-$ SA2++ & 95\% interval & bf16 spread \\
\midrule
Self-Forcing, 11\,s & 946 & $-0.17$ & $-19.2$ & $-0.25$ & $-0.08$ & [$-0.16$, $+0.01$] & 0.23 \\
Self-Forcing, 60\,s & 128 & $-0.60$ & $-27.9$ & $+0.02$ & $+0.62$ & [$-1.33$, $+2.34$] & 3.12 \\
Causal Forcing & 32 & $-0.32$ & $-15.6$ & $-0.57$ & $\mathbf{-0.25}$ & [$-0.51$, $-0.01$] & 0.32 \\
CausVid & 32 & $-0.06$ & $-9.2$ & $-0.25$ & $\mathbf{-0.19}$ & [$-0.29$, $-0.08$] & 0.06 \\
Context Forcing & 32 & $-0.38$ & $-13.4$ & $-0.56$ & $-0.17$ & [$-1.32$, $+0.91$] & 1.70 \\
LongLive-1.3B & 32 & $-0.07$ & $-16.4$ & $-0.17$ & $-0.10$ & [$-0.71$, $+0.44$] & 0.52 \\
LongLive-2.0-5B & 32 & $+0.16$ & $-9.4$ & $-0.23$ & $-0.39$ & [$-1.54$, $+0.73$] & 1.60 \\
LingBot-World v2 & 32 & $+0.18$ & $-13.3$ & $-0.48$ & $\mathbf{-0.66}$ & [$-1.09$, $-0.23$] & 0.85 \\
ABot-World-0 & 32 & $-0.04$ & $-3.8$ & $-0.27$ & $-0.23$ & [$-0.68$, $+0.22$] & 0.82 \\
\bottomrule
\end{tabular}
\par\vspace{2pt}
{\footnotesize $^\ast$Evaluated on 128 and 32 prompts (Self-Forcing) and 8 prompts (Causal Forcing, CausVid, Context Forcing, and both LongLive models). Details: \cref{app:models}.\par}
\end{table*}

\begin{table*}[t]
\caption{\textbf{VBench-Long scores on LongLive-2.0-5B} (LongLive~2.0's protocol: 60-s videos of MovieGen prompts, bf16 weights, whole video, score $\times100$, higher is better). Perturbed: bf16 with 0.17\% attention-output noise (\cref{app:protocol}). Flagged: dimensions of 6 significantly separated from bf16. \sagethree{} is evaluated on 8 prompts; the last two columns report means over those 8. Dynamic degree is the share of 2-s clips classified as dynamic (paired standard error $\approx 3$ points on 32 prompts).}
\label{tab:vblong}
\centering
\small
\setlength{\tabcolsep}{4pt}
\begin{tabular}{@{}lccccccc@{}}
\toprule
 & \multicolumn{5}{c}{32 prompts} & \multicolumn{2}{c}{8 prompts} \\
\cmidrule(lr){2-6}\cmidrule(l){7-8}
 & bf16 & perturbed & \sagetwo{} & \kvq{} & \textbf{\ours{}} & bf16 & \sagethree{} \\
\midrule
Subject consistency & 98.05 & 97.91 & 98.01 & 97.81 & 98.10 & 97.97 & 97.50 \\
Background consistency & 97.19 & 97.14 & 97.22 & 97.07 & 97.22 & 97.38 & 97.08 \\
Motion smoothness & 99.07 & 98.99 & 99.03 & 99.03 & 99.09 & 99.25 & 99.21 \\
Dynamic degree & 56.25 & 56.04 & 50.62 & 55.42 & 52.19 & 50.42 & 45.42 \\
Aesthetic quality & 58.67 & 59.13 & 58.06 & 58.35 & 59.16 & 60.87 & 59.77 \\
Imaging quality & 68.14 & 68.32 & 67.98 & 68.46 & 68.52 & 69.11 & 65.82 \\
Flagged & --- & --- & 0/6 & 0/6 & 0/6 & --- & 3/6 \\
\bottomrule
\end{tabular}
\end{table*}

\begin{figure*}[t]
  \centering
  \safefig{\textwidth}{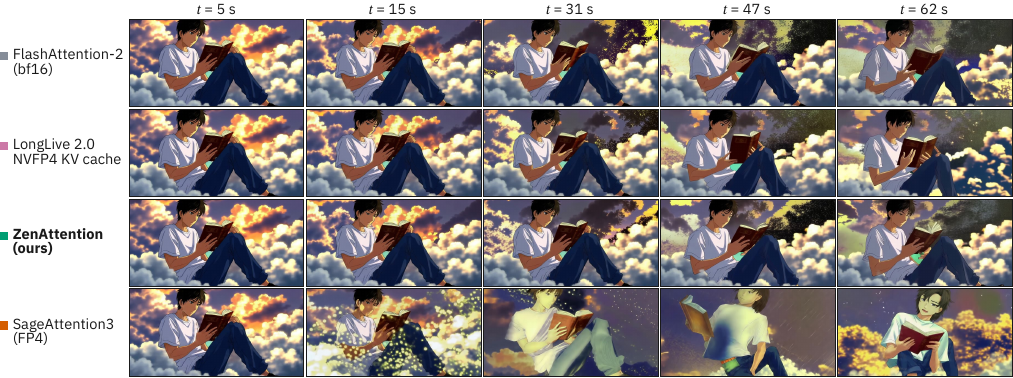}
  \caption{\textbf{A one-minute rollout} (LongLive-2.0-5B, ``reading on a cloud'', $704\times1280$ at half resolution; one rollout per row, same prompt and seed). \sagethree{} breaks into specks by 15\,s, then changes the character and style; \ours{} stays with bf16, as does LongLive~2.0's NVFP4 cache, which is as slow as bf16. The other paths of this rollout and a second scene are in \cref{fig:qualread,fig:qualgold}.}
  \label{fig:qualtime}
\end{figure*}

\subsection{Setup}
\label{sec:setup}

\textbf{Models.} We evaluate eight streaming models: Self-Forcing \citep{huang2025selfforcing}, Causal Forcing \citep{zhu2026causalforcing}, Context Forcing \citep{chen2026contextforcing}, CausVid \citep{yin2025causvid}, LongLive-1.3B \citep{yang2026longlive} and the camera-controlled world model LingBot-World v2 1.3B \citep{robbyant2026lingbotworld}, all based on Wan~2.1 1.3B \citep{wan2025wan}; and LongLive-2.0-5B \citep{chen2026longlive2} and the interactive world model ABot-World-0 \citep{jiang2026abotworld}, both based on Wan~2.2 5B at $704\times1280$. \ours{} runs on all of their cache layouts without retraining (\cref{app:models}). All runs use batch size 1 and each model's default settings.

\textbf{Baselines.} FlashAttention-2 in bf16 \citep{dao2024flashattention2}; \sagetwo{} \citep{zhang2025sageattention2pp} and \sagethree{} \citep{zhang2025sageattention3}; the persistent INT8/FP8 cache of LightX2V \citep{lightx2v2025lightx2v} with its own kernels; and the NVFP4 cache of LongLive~2.0 \citep{chen2026longlive2} with its own compiled quantizer and inference script.

\textbf{Metrics and timing.} We report VBench \citep{huang2024vbench} on VBench prompts at 11\,s and MovieGen prompts \citep{polyak2024moviegen} at 60\,s; the latent detail metric of \cref{sec:analysis}; and effective TOPS, $4N_qN_kdH/t$, where $t$ is the wall time of one call including every launch it needs (our query preparation and the per-call quantization of \sagetwo{} and \sagethree{}). Self-attention time per video sums all attention calls and cache writes; denoising times are medians of five runs of each model's official inference script. All speed measurements are collected on an NVIDIA RTX~5090. In quality evaluations the first chunk uses the reference kernel (\cref{app:protocol}).

\subsection{Speed}
\label{sec:speed}

\textbf{Kernel.} On one call with 23,400 keys (\cref{fig:speed}, left), \ours{} reaches 628 effective TOPS against 433 for \sagetwo{}, 432 for \sagethree{} and 183 for FlashAttention-2; with 32,760 keys it reaches 637 against 430 ($1.48\times$; \cref{fig:kernel} in \cref{app:kernel}). Its advantage per call grows with the cached history, because only the FP4 tier grows. The persistent 8-bit cache of LightX2V is within $-3$ to $+5\%$ of \sagetwo{}, so quantizing once gains nothing without FP4 products, and \sagethree{}'s FP4 products gain nothing when the whole window is requantized at every call. \Cref{sec:kernel} accounts for the gap between these numbers and the $1.98\times$ of pure tensor-core loops.

\textbf{Self-attention.} In full rollouts (\cref{fig:speed}, right; \cref{tab:speed}), \ours{} runs self-attention $2.6$--$3.5\times$ faster than bf16 and $1.23$--$1.50\times$ faster than \sagetwo{} across all eight models. \sagethree{} is slower than \sagetwo{} on all models except LongLive-2.0-5B. On Self-Forcing, \ours{} is also $1.34\times$ faster than the LightX2V cache in attention time without cache writes.

\textbf{Denoising.} In each model's official pipeline, denoising is $1.21\times$ faster than with \sagetwo{} on Self-Forcing (52.2 vs.\ 63.0\,s for 60\,s of video; 98.8\,s with bf16, $1.89\times$), $1.14\times$ on LongLive-1.3B (60.7 vs.\ 69.3\,s; 87.8\,s with bf16) and $1.12\times$ on LongLive-2.0-5B (\cref{fig:speed}, middle). The gain grows with video length as the cache fills, from $1.10\times$ at 4\,s to $1.21\times$ at 60\,s on Self-Forcing. With LongLive~2.0's W4A4 linear layers, which run as native FP4 products, \ours{} replaces the release's NVFP4 cache and denoises 64\,s of video in 68.8 instead of 126.4\,s ($1.84\times$) at the same peak memory (within 0.1\,GiB).

\textbf{Where the denoising gain comes from.} The write-once design saves time twice: the kernel is faster, and there is no bf16 window left to maintain, which a plug-in kernel such as \sagetwo{} must keep. Of the 13.3\,s saved on Self-Forcing, 8.8\,s are kernel time and 3.4\,s come from eliminating bf16 window maintenance (\cref{app:models}).

\begin{table}[t]
\caption{\textbf{Ablation} on Self-Forcing (18 latent frames, 8 held-out scenes, chunks 1--5): change of latent detail and latent norm (\%) against exact attention; reference spread 0.03. Emulated FP4 tier in fp32.}
\label{tab:ablation}
\centering
\small
\setlength{\tabcolsep}{4pt}
\begin{tabular}{lrr}
\toprule
Configuration & $\Delta$detail & $\Delta$norm \\
\midrule
Exact attention (2nd run) & $+0.00$ & $+0.01$ \\
Plain NVFP4, no correction & $-4.41$ & $-3.74$ \\
\quad only $QK^{\top}$ in FP4 & $-3.86$ & $-1.81$ \\
\quad only $\widetilde PV$ in FP4 & $-0.27$ & $-1.64$ \\
\rowcolor{tierfour!10}
\textbf{\ours{}} & $\mathbf{-0.07}$ & $\mathbf{-0.07}$ \\
\quad w/o scale correction & $-1.02$ & $-0.55$ \\
\quad w/o variance correction & $-0.47$ & $-0.17$ \\
\quad round-to-nearest $\widetilde P$ (no dither) & $+0.65$ & $+0.09$ \\
\quad unrounded denominator & $-0.04$ & $-0.22$ \\
\quad w/o mean/residual split & $-0.01$ & $-0.11$ \\
\bottomrule
\end{tabular}
\end{table}

\subsection{Quality and Memory}
\label{sec:e2e}

\textbf{VBench.} On Self-Forcing (\cref{tab:vbench}), no cell is flagged for \ours{} or \sagetwo{} on all 946 VBench prompts at 11\,s or on 128 MovieGen prompts at 60\,s. At 946 prompts, every 95\% bootstrap interval of \ours{} minus bf16 lies within $\pm0.09$ points. By contrast, \sagethree{} is flagged in 10 of 21 cells at 11\,s and 14 of 21 at 60\,s, with losses up to 5 and 19 points. On LongLive-2.0-5B under LongLive~2.0's VBench-Long protocol (32 MovieGen prompts, 60\,s; \cref{tab:vblong}), no dimension separates \ours{} from bf16 or \sagetwo{}, whereas \sagethree{} is flagged in 3 of 6. In bootstrap 95\% intervals over prompts, every difference between \ours{} and bf16 includes zero, while \ours{} outperforms LongLive~2.0's own NVFP4 cache by $0.29$ points in subject consistency and $0.14$ in background consistency (\cref{app:longlive2}).

\textbf{Latent detail.} \Cref{tab:models} measures latent detail across eight models, a finer scale than VBench. \ours{} stays within $0.7\%$ of bf16 on every model, matching \sagetwo{}, whereas \sagethree{} loses 4--28\% and visibly degrades the scene (\cref{fig:qualtime}). LongLive~2.0's own NVFP4 cache, which computes in bf16, is at $-0.48$ points on LongLive-2.0-5B (interval [$-1.40$, $+0.44$]). Comparing \ours{} directly with 8-bit attention, the two stay within $0.7$ points on every model, and on five of eight---including all 946 Self-Forcing prompts---the 95\% interval of their difference includes zero (\cref{app:models}).

\textbf{Stability over one minute.} One-minute Self-Forcing rollouts occasionally turn hazy after the window starts to slide, with any kernel including bf16. On 128 prompts, detail in chunks 10--14 falls more than 3 points below the reference in 11 rollouts of a second bf16 run, 11 of \sagetwo{}, and 14 of \ours{}; the rates do not differ in paired tests ($p=0.65$ and $0.66$, \cref{app:drops}).

\textbf{Memory.} The records take $33$--$35\%$ of the bytes of a bf16 cache and $69\%$ of an 8-bit cache (177, 258 and 512 bytes per key and head on Self-Forcing, \cref{tab:options}). The allocated Self-Forcing cache takes 1.94\,GiB, against 2.83\,GiB for the LightX2V cache and 5.62\,GiB in bf16; on LongLive-2.0-5B the records take 3.23\,GiB against 9.67\,GiB in bf16. Since the kernel does not read a bf16 cache, none needs to be allocated: peak denoising memory falls from 18.6 to 16.4\,GiB on LongLive-1.3B (18.7\,GiB with \sagetwo{}, which keeps the bf16 cache) and, with bf16 weights, from 24.3 to 17.5\,GiB on LongLive-2.0-5B. This matches the peak of LongLive~2.0's own NVFP4 cache (17.4\,GiB), which brings no speed-up; with LongLive~2.0's W4A4 linear layers both peak at 11\,GiB, and \ours{} is $1.84\times$ faster (\cref{app:longlive2}).

\subsection{Ablation Study}
\label{sec:ablation}

\Cref{tab:ablation} removes one component at a time. The scale correction is the largest single contributor: without it, detail drops to $-1.02\%$ and norm to $-0.55\%$ across all 8 scenes. Removing the variance correction degrades detail to $-0.47\%$ (8 of 8 scenes), while rounding $\widetilde P$ to nearest instead of dithering flips the error sign and oversharpens the video ($+0.65\%$, 8 of 8 scenes). Using the unrounded softmax denominator shrinks the output norm ($-0.22\%$). By contrast, removing the mean/residual split of keys and queries leaves detail at $-0.01\%$ and norm at $-0.11\%$ (within the noise floor, with signs split 4--4 across scenes). This confirms that \ours{}'s recovery is driven by our bias corrections rather than standard $Q/K$ smoothing; we retain the mean/residual split because its terms fold into the precomputed vector $C$ with no extra inner-loop FMAs and safeguard channels with large mean offsets. Finally, removing the 8-bit tier for the newest frame drops first-frame contrast by $0.42$ points (\cref{sec:copyhead}). The variance correction, the dither, and the dominance of the score side differ from observations in other settings (\cref{sec:related}).

\textbf{8-bit attention with 4-bit linear layers.} LongLive~2.0 ships its W4A4 linear layers with bf16 attention, and SageAttention has no W4A4 configuration of its own. Combining \sagetwo{} with LongLive~2.0's W4A4 linear layers denoises 64\,s of video in 82.6\,s versus 70.2\,s for \ours{} ($1.18\times$), while requiring 18.1 vs.\ 11.0\,GiB of peak memory (\cref{app:longlive2}).

\section{Related Work}
\label{sec:related}

\paragraph{Low-bit attention.} SageAttention quantizes attention operands for tensor cores at 8 and 4 bits \citep{zhang2025sageattention,zhang2025sageattention2,zhang2025sageattention2pp,zhang2025sageattention3} on top of the FlashAttention online softmax \citep{dao2022flashattention,dao2024flashattention2,shah2024flashattention3}. VC-Attention smooths the values for 4-bit attention in bidirectional video models \citep{li2026vcattention}, and DraftAttention2 mixes 4-, 8- and 16-bit phases in one kernel \citep{ding2026draftattention2}. These quantize their operands at every call.

\paragraph{Low-bit caches for streaming video.} Caches of autoregressive video and world models have been compressed to 2--4 bits \citep{ranganath2026kvcache,xi2026quantvideogen,lucas2026trq,zhao2026quantwm,hu2026tempokv,ge2026coralkv,tuncer2026quantizedkeys,chen2026longlive2}; all dequantize before attention and save memory, not time. A persistent 8-bit cache read directly by an 8-bit kernel exists in LightX2V \citep{lightx2v2025lightx2v}. \citet{tuncer2026quantizedkeys} identify the shift of mass towards noisy quantized keys and correct it for exact queries. Compounding of cache errors has been noted \citep{ge2026coralkv,guo2026laq}; we measure it across chunks. Heads attending to the most recent frame are used for pruning \citep{ji2026forcingkv,tian2026headforcing,shen2026headcast}, not for precision.

\paragraph{Computing on low-bit caches in language models.} In language-model decoding, BitDecoding dequantizes a packed low-bit cache in registers \citep{du2026bitdecoding}, and BitDecoding-FP4, NVFP4-MLA and MpFA multiply stored FP4 keys natively \citep{du2026bitdecodingfp4,ni2026nvfp4mla,deng2026mpfa}; KIVI keeps recent tokens at higher precision \citep{liu2024kivi}. Closest to our design, ScaleSearchAttention \citep{gupta2026scalesearch} stores completed key and value blocks once in NVFP4 and keeps the first block and the incomplete most recent block at full precision; it is evaluated in a PyTorch simulation on language models, without a kernel for this path. Our setting differs in that a compiled kernel serves thousands of queries per call, the cache receives the model's own output, and the bias is corrected. To our knowledge no prior work computes attention directly on a stored 4-bit cache for autoregressive video diffusion.

\paragraph{Bias correction.} Debiasing quantized inner products appears in DRIVE, EDEN and RaBitQ \citep{vargaftik2021drive,vargaftik2022eden,gao2024rabitq}, per-channel correction in \citet{banner2019posttraining}, softmax bias correction in \citet{pandey2023softmax}, and the rounded-sum denominator in \citet{yu2026mxattention,anthony2026zyfa}; stochastic rounding is classical \citep{gupta2015limited,croci2022stochastic}. Unlike a persistent offset, which harms planners \citep{song2026dithered}, our dither varies with query, key block and head. Our contribution is their combination for a write-once cache: the statistics are stored with the records, and the corrections cost one fused multiply-add per score.

\paragraph{Random rotation.} A random Hadamard rotation makes quantization errors nearly input-independent and is standard in 4-bit inference \citep{ashkboos2024quarot}; DRIVE and EDEN rely on it for their unbiasedness guarantees. We omit random rotations because 16-element microscaling blocks already adapt locally without adding per-call query rotation overhead, and validate the resulting corrections directly (\cref{tab:ablation,fig:compounding}).

\paragraph{Differences from prior observations.} Three of our measurements differ from reports in other settings. Without the variance correction the loss of detail grows from $0.07$ to $0.47$ points, so under NVFP4 with quantized queries the Jensen bias is not mild, unlike the INT4 setting with exact queries of \citet{tuncer2026quantizedkeys}. Dithered rounding of $\widetilde P$ beats rounding to nearest, whereas \citet{chatelain2026stochastic} find stochastic rounding worse at 4 bits; they round accumulated products rather than the operand $\widetilde P$. In streaming rollouts the score side dominates the error, whereas \citet{li2026vcattention} attribute most of the error of 4-bit attention in bidirectional video models to the values.

\section{Limitations}
\label{sec:limits}

\ours{} pays off when attention over a multi-frame cache accounts for a substantial share of denoising time; with a very short cache, or when attention is a small share of the model, the gain is smaller (\cref{app:models}). Against 8-bit attention it gives up a small amount of latent detail on some models ($0.2$--$0.7$ points on three of eight). Our controlled error analysis focuses on Self-Forcing, and the compiled kernel targets NVIDIA Blackwell GPUs (compute capability 12.0) with head dimension 128.

\section{Conclusion}
\label{sec:conclusion}

We set out to keep both the KV cache and the attention computation of streaming video diffusion at 4 bits. Doing this naively fails for two reasons: the error of FP4 attention is biased and written back into the cache, amplifying single-chunk detail loss roughly $3\times$ on Self-Forcing; and the newest cached frame needs more precise scores than older history, traced in Wan~2.1 to a single copy head. \ours{} counters the bias in closed form and keeps the newest cached frame at 8 bits. Across eight streaming video and world models without retraining, the compiled kernel reaches 628--637 effective TOPS on an RTX~5090, accelerates self-attention by $2.6$--$3.5\times$ over bf16 ($1.23$--$1.50\times$ over \sagetwo{}, beating both 8-bit and 4-bit SageAttention), and speeds up end-to-end denoising by up to $1.89\times$ over bf16 ($1.84\times$ over LongLive~2.0's 4-bit release). It cuts KV-cache memory by two-thirds while keeping latent detail within $0.7\%$ of bf16 (vs.\ 4--28\% lost by \sagethree{}) and matching bf16 on VBench for Self-Forcing and LongLive-2.0-5B.

\section*{Use of Generative AI}
We used generative AI tools (large-language-model coding assistants) to assist with experimental design, with writing and debugging experiment and analysis code, with running experiments and analysing their results, with searching the related literature, and with drafting and editing the manuscript. All AI-assisted outputs, including code, numerical results, citations and text, were reviewed and verified by the authors, who take full responsibility for the content of the paper.

\bibliographystyle{abbrvnat}
\bibliography{refs}

\clearpage
\onecolumn
\appendix

\section{Score and variance derivation}
\label{app:derivation}

\paragraph{Rescaled score and variance.} Write the rescaled residuals as $\hat{\tilde q}_i/\alpha_i=\tilde q_i+u_i$ and $\beta_j\hat r_j=r_j+w_j$; by the definitions of $\alpha_i$ and $\beta_j$, $\langle u_i,\tilde q_i\rangle=0$ and $\langle w_j,r_j\rangle=0$. Substituting $q_i=\tilde q_i+\bar q_g$ and $k_j=r_j+\bar k_c$ into the score $\tilde s_{ij}$ of \cref{eq:score} before subtracting $\tfrac12\sigma^2_{ij}$ gives $\sqrt d\,(\tilde s_{ij}-s^{\star}_{ij})=\langle u_i,r_j\rangle+\langle q_i,w_j\rangle+\langle u_i,w_j\rangle$, where $s^{\star}_{ij}=\langle q_i,k_j\rangle/\sqrt d$. Under an idealized error model in which $u_i$ and $w_j$ are independent, zero-mean, and isotropic in $\mathbb R^d$, $\mathbb E\,\tilde s_{ij}=s^{\star}_{ij}$ and the leading-order score variance is $(\|u_i\|^2\|r_j\|^2+\|q_i\|^2\|w_j\|^2)/d^2$ (the cross term $\langle u_i,w_j\rangle$ is higher order in $1/d$); for a Gaussian score error of variance $\sigma^2_{ij}$, $\mathbb E\,e^{\tilde s_{ij}-\sigma^2_{ij}/2}=e^{s^{\star}_{ij}}$. \Cref{eq:score} evaluates $\sigma^2_{ij}$ with the realized pre-rescaling norms $\|\hat{\tilde q}_i-\tilde q_i\|^2$, $a_j=\|\hat r_j\|^2$, and $e_j=\|\hat r_j-r_j\|^2$ (which differ by $\alpha_i,\beta_j\approx1$), and the kernel averages the query norms over blocks of 64 queries (\cref{sec:kernel}).

\paragraph{Dithered rounding and approximations.} Below 2 code units, the positive E2M1 grid $\{0,\tfrac12,1,\tfrac32,2\}$ has uniform step $\Delta=\tfrac12$, so adding $D\sim\mathcal U[-\tfrac14,\tfrac14)$ before nearest-neighbor rounding yields standard unbiased stochastic rounding on $[0,2)$ (averaged over positions of the fixed $64\times32$ lookup table in the kernel; weights $\ge2$ code units fall on coarser E2M1 steps). Because NVFP4 rounding without random rotations is deterministic and each output row divides by a rounded denominator sum, these unbiasedness identities hold only approximately and are validated empirically in \cref{tab:ablation,fig:compounding}.

\section{Evaluation protocol}
\label{app:protocol}

\paragraph{References.} The primary reference is the bf16 baseline of each model's official pipeline; the \emph{noise-floor run} is the same pipeline with an independent $0.17\%$ relative perturbation of every attention output. For the mechanism and compounding studies the reference is exact fp32 attention with the same perturbation. Every run computes the first chunk with the reference kernel so trajectories start identically. We compare compiled kernels with compiled references and emulations with emulated references.

\paragraph{Decision criteria.}
(i) \emph{Compounding} (\cref{sec:compound}): a scene exhibits compounding if (1) the per-call error of the uncorrected kernel does not grow across chunks; (2) its free detail loss grows from chunks 1--4 to chunks 10--14 by more than the reference spread; and (3) its compounded part (free minus teacher-forced deviation, chunks 5--14) exceeds the reference spread.
(ii) \emph{VBench} (\cref{sec:e2e}): a cell is flagged if the mean paired difference to bf16 exceeds three times the larger of the noise-floor run's mean and standard error, and at least $\lceil 7n/8\rceil$ of $n$ prompts share its sign. To bound smaller shifts as well, we complement this test with prompt-level 95\% bootstrap intervals (all within $\pm0.09$ points at $n=946$).
(iii) \emph{Per-prompt divergence} (\cref{sec:e2e}): a prompt diverges if \ours{} deviates from the bf16 mean by at least 3 points of detail while \sagetwo{} stays within 2.

\paragraph{Sample sizes.} VBench is evaluated on all 946 VBench prompts at 11\,s (\sagethree{} and the LightX2V cache on 128 prompts) and 128 MovieGen prompts at 60\,s (\sagethree{} and LightX2V on 32 prompts). Within each of the 11 VBench prompt categories (72--100 prompts each), no cell is flagged for \ours{} or \sagetwo{}. Confidence intervals use 10{,}000 bootstrap resamples over prompts. Latent detail sample sizes are listed in \cref{tab:models}; mechanism ablations and compounding experiments use 8 held-out Self-Forcing scenes.

\section{Sign and size of the compounded error}
\label{app:controls}

The corrections remove the signed part of the per-call error and halve its size (\cref{sec:compound}). To see which of the two the cache carries forward, we inject errors of a chosen sign structure and magnitude, in the protocol of \cref{sec:compound} (same 8 scenes, exact base rollouts and noise-floor runs; one free and one teacher-forced rollout per error and scene).

\paragraph{Injected errors.} At every attention call from chunk 1 on we compute the exact output $E$ and the output $X$ of the uncorrected FP4 operator on the same inputs, and emit $E$ plus an error built from $d=X-E$: the \emph{negated} error, $E-d$; the \emph{sign-randomized} error, $E+s\odot d$ with independent $s_i=\pm1$, which keeps the size of every element and has zero mean; and zero-mean \emph{Gaussian} noise, $E\odot(1+\epsilon\,n)$ with $n\sim\mathcal N(0,1)$ per element, where $\epsilon$ is the uncorrected operator's mean relative error in that layer and denoising pass ($2$--$36\%$ across layers).
Emitting $X$ itself through this code path reproduces the uncorrected rollout of \cref{sec:compound} bit for bit. In the free rollouts the negated and the sign-randomized errors measure $7.8\%$ and $7.7\%$ per call; the signed gain of the output along the exact output, $-1.47\%$ for the uncorrected operator ($-1.59\%$ in the high-pass band), is $+1.47\%$ for the negated error and $0.00\%$ for the sign-randomized one.

\begin{table}[h]
\caption{\textbf{Sign and size controls} on Self-Forcing (45 latent frames, 8 held-out scenes; means over scenes). Error: relative $L_2$ error of one attention call (\%). One chunk: teacher-forced detail deviation, chunks 5--14 (\%). Free: detail deviation of the free rollout, chunks 10--14 (\%). Compounded: teacher-forced minus free deviation, chunks 5--14 (points; positive: the free rollout lost more).}
\label{tab:controls}
\centering
\small
\setlength{\tabcolsep}{4pt}
\begin{tabular}{lrrrr}
\toprule
Attention output & Error & One chunk & Free & Compounded \\
\midrule
uncorrected operator, $E+d$ & $7.5$ & $-3.45$ & $-10.27$ & $+6.10$ \\
negated, $E-d$ & $7.8$ & $+4.63$ & $+17.87$ & $-10.37$ \\
signs randomized, $E+s\odot d$ & $7.7$ & $-0.15$ & $-3.20$ & $+2.35$ \\
Gaussian, $1\times$ & $7.6$ & $-0.10$ & $-3.20$ & $+2.48$ \\
two gains per head only & $2.0$ & $-4.69$ & $-19.19$ & $+11.72$ \\
remainder, rescaled to $\|d\|$ & $7.6$ & $+1.33$ & $+8.64$ & $-4.85$ \\
\midrule
Gaussian, $0.5\times$ & $3.8$ & $-0.03$ & $-0.78$ & $+0.65$ \\
Gaussian, $1.5\times$ & $11.4$ & $-0.24$ & $-5.00$ & $+3.72$ \\
\midrule
\sagetwo{} & $1.7$ & $-0.10$ & $-1.46$ & $+0.86$ \\
\ours{} & $3.9$ & $-0.03$ & $+0.14$ & $+0.05$ \\
\sagethree{} & $11.3$ & $-8.54$ & $-25.29$ & $+15.44$ \\
\bottomrule
\end{tabular}
\end{table}

\begin{figure}[h]
  \centering
  \safefig{\textwidth}{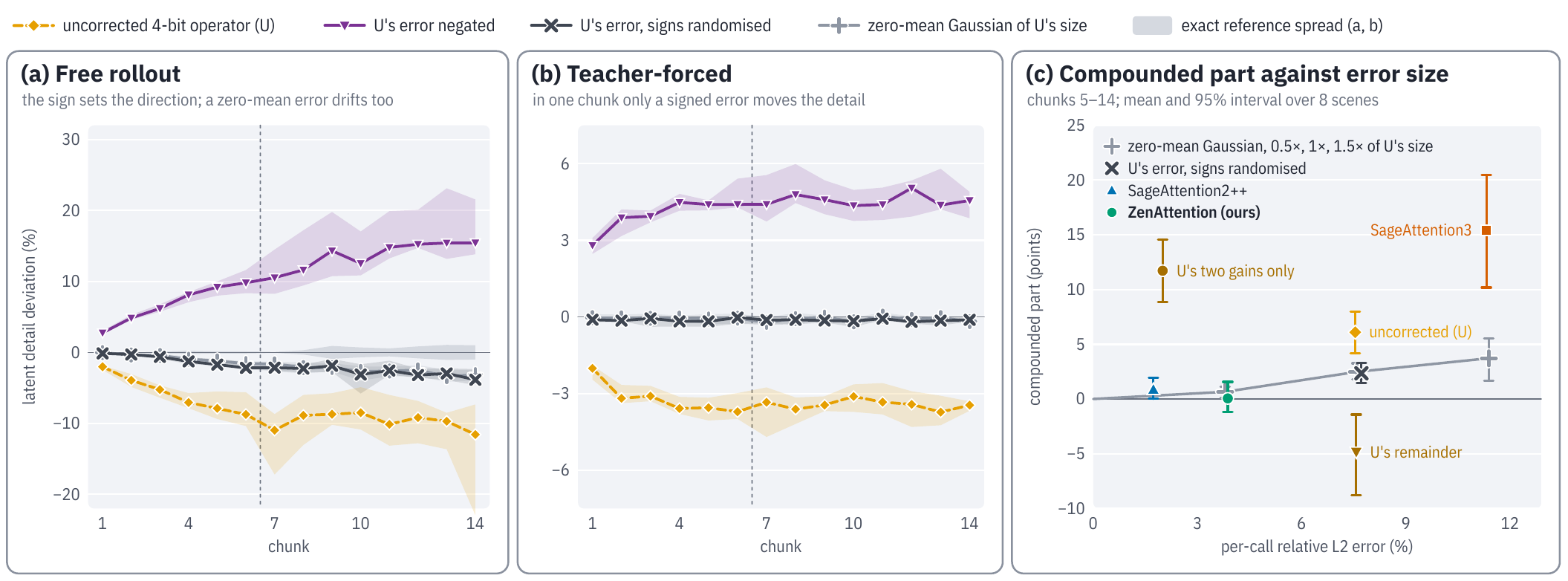}
  \caption{\textbf{Sign and size of the compounded error} (Self-Forcing, 8 held-out scenes; \cref{app:controls}). (a, b)~Free and teacher-forced detail deviation for errors built from the uncorrected operator's own error; medians over scenes with interquartile bands. (c)~Compounded part against the per-call error, means with bootstrap 95\% intervals over scenes; the grey line joins the zero-mean Gaussian errors of three sizes, and the two parts of the uncorrected operator's error are labeled.}
  \label{fig:controls}
\end{figure}

\paragraph{What the controls show.} \Cref{tab:controls} and \cref{fig:controls} show three clear effects:
(i) \emph{Single-chunk loss comes from the signed part.} With an exact cache, the two zero-mean errors change detail by only $-0.15\%$ and $-0.10\%$, against $-3.45\%$ for the uncorrected operator, while the negated error raises detail by $4.63\%$.
(ii) \emph{Rollout drift follows the sign.} In free rollouts the negated error drives detail up to $+17.9\%$ in chunks 10--14, whereas the uncorrected operator loses $10.3\%$; the odd (sign-reversing) component of the compounded error is $O=(\bar K_U-\bar K_N)/2 = 8.24$ points (above the $1.06$ reference spread in all 8 scenes), compared to $V=(\bar K_U+\bar K_N)/2 = -2.14$ points.
(iii) \emph{Magnitude also compounds, at a smaller scale.} The two zero-mean errors lose almost nothing in one chunk yet lose $3.2\%$ in free rollouts (compounded parts $2.35$ and $2.48$ points, about $0.4\times$ the uncorrected operator's $6.10$).

\paragraph{Size sweep.} Scaling zero-mean Gaussian error to $0.5\times$ and $1.5\times$ (matching the per-call error sizes of \ours{} and \sagethree{}) yields compounded parts of $0.65$, $2.48$, and $3.72$ points. Halving a zero-mean error removes $1.82$ points [$1.42$, $2.35$], whereas \ours{}'s corrections remove $6.05$ points [$3.92$, $8.20$] (paired difference $4.23$ [$2.24$, $6.28$]). Relative to a zero-mean error of the same magnitude, the uncorrected operator compounds $3.65$ points more [$1.57$, $5.70$] and \sagethree{} $11.74$ points more [$6.15$, $18.00$], whereas \ours{} ($-0.64$ [$-1.96$, $0.76$]) and \sagetwo{} ($+0.56$ [$-0.37$, $1.77$]) stay indistinguishable from the zero-mean curve.

\paragraph{Gain decomposition.} Splitting $d$ per head and call into its least-squares projection onto the low-pass and high-pass components of $E$ (two signed gains per head) and the orthogonal remainder (rescaled to $\|d\|$) further highlights the dominance of signed bias: the two gains alone carry only $2.0\%$ per-call error yet compound to $11.72$ points [$8.88$, $14.64$], nearly double the full uncorrected operator ($6.10$), while the $7.6\%$ remainder pushes detail in the opposite direction ($+1.33\%$ in one chunk, $+8.6\%$ in free rollouts, compounded part $-4.85$ points). Thus both signed bias and error size contribute to compounding, with signed bias carrying the dominant effect.

\paragraph{Corrections switched inside the emulation.} The uncorrected operator and all injected errors are emulated in fp32, whereas \ours{} above is the compiled kernel. Switching only the correction flags and the 8-bit newest-frame tier inside the same fp32 emulation lowers the mean compounded part across all 8 scenes from $6.10$ to $0.23$ points (paired reduction $5.87$, bootstrap 95\% interval [$3.67$, $8.22$]; lower in 8 of 8 scenes).

\section{Models and per-model results}
\label{app:models}

\begin{table}[h]
\caption{\textbf{Time per rollout} (s, RTX~5090) for self-attention, including cache writes, and for the whole denoising network in each model's official pipeline. Speed-up reports \ours{} over \sagetwo{}. W4A4: LongLive~2.0's 4-bit release (W4A4 linear layers with its NVFP4 KV cache) against the same linear layers with \ours{}.}
\label{tab:speed}
\centering
\small
\setlength{\tabcolsep}{4pt}
\begin{tabular}{@{}lrrrrr@{}}
\toprule
Model (video) & bf16 & SA3 & SA2++ & \textbf{Ours} & Speed-up \\
\midrule
\multicolumn{6}{@{}l}{\emph{Self-attention}} \\
Self-Forcing (60\,s) & 56.9 & 29.2 & 25.37 & \textbf{17.32} & $1.46\times$ \\
Causal Forcing (60\,s) & 57.49 & 28.36 & 24.43 & \textbf{16.49} & $1.48\times$ \\
Context Forcing (60\,s) & 57.49 & 28.32 & 24.16 & \textbf{17.04} & $1.42\times$ \\
CausVid (5\,s) & 2.43 & 1.58 & 1.25 & \textbf{0.83} & $1.50\times$ \\
LongLive-1.3B (60\,s) & 33.69 & 16.99 & 14.56 & \textbf{10.82} & $1.35\times$ \\
LingBot-World v2 (17--22 chunks)$^\ddagger$ & 15.85 & 7.34 & 6.53 & \textbf{5.32} & $1.23\times$ \\
LongLive-2.0-5B (128\,s) & 159.37 & 65.03 & 68.78 & \textbf{55.57} & $1.24\times$ \\
ABot-World-0 (60 chunks)$^\ddagger$ & 25.86 & 18.93 & 13.47 & \textbf{9.81} & $1.37\times$ \\
\midrule
\multicolumn{6}{@{}l}{\emph{Denoising}} \\
Self-Forcing (60\,s) & 98.82 & --- & 62.97 & \textbf{52.17} & $1.21\times$ \\
LongLive-1.3B (60\,s) & 87.84 & --- & 69.32 & \textbf{60.69} & $1.14\times$ \\
LongLive-2.0-5B (64\,s) & 164.55 & --- & 119.77 & \textbf{107.25} & $1.12\times$ \\
\quad W4A4 & \multicolumn{3}{r}{4-bit release: 126.38} & \textbf{68.77} & $1.84\times$ \\
\bottomrule
\end{tabular}
\\[2pt]{\footnotesize Speed-ups computed from unrounded times. $^\ddagger$Mean over 32 settings. LongLive~2.0's NVFP4 cache with bf16 weights takes 164.83\,s denoising. With bf16 attention, self-attention accounts for $38\%$ of denoising on LongLive-1.3B, $48\%$ on LongLive-2.0-5B (78.7\,s, \cref{app:longlive2}), and $58\%$ on Self-Forcing.}
\end{table}

\paragraph{Details of \cref{tab:models}.} Videos are 5\,s long for Causal Forcing and CausVid, 60\,s for Context Forcing and LongLive-1.3B (last 30\,s), 17--22 chunks for LingBot-World, and 60 chunks for ABot-World-0; LongLive-2.0-5B covers chunks 8--23 of the 60-s rollouts on the 32 MovieGen prompts (\cref{app:longlive2}). Confidence intervals are bootstrap intervals over prompts for video models and $t$-intervals for the world models. On the 946 Self-Forcing prompts, \ours{} is $0.15\pm0.01$ points below \sagetwo{} in the first third of the video and $0.08$ below over the whole video. On LingBot-World the paired difference is $1.2\pm0.4$ points in the last third; on Causal Forcing, CausVid, and Context Forcing, per-prompt divergence exceeds 3 points on at most 1 of 32 prompts over the whole video.

\paragraph{Where the denoising gain comes from.} Self-attention times in \cref{tab:speed} use a shared cache harness, whereas denoising times use each model's official script (where \sagetwo{} retains the model's bf16 window). Synchronized profiling splits the gain over \sagetwo{} into kernel time (net of cache writes) and elimination of bf16 window maintenance: on Self-Forcing (60\,s), 8.8\,s from the kernel and 3.4\,s from the window (out of 13.3\,s); on LongLive-1.3B (60\,s), 4.2\,s and 4.2\,s (out of 8.7\,s); on LongLive-2.0-5B (64\,s), 7.3\,s and 5.9\,s (out of 12.9\,s), where the official pipeline clones the bf16 window on every self-attention call; and with W4A4 linear layers, 7.0\,s and 5.6\,s (out of 11.6\,s).

\paragraph{Scope of applicability.} SANA-Video uses linear attention, SkyReels-V2 and bidirectional Wan keep no KV cache, NOVA has head dimension 64 and reads its cache once per frame, and Rolling Forcing denoises five chunks jointly (23{,}400 queries per call, so only about a quarter of the keys are in the 4-bit tier). Context Forcing re-applies positional rotations when its context shifts, requiring partial cache updates (about four group writes per chunk, $4\%$ of attention time).

\section{LongLive~2.0 comparison details}
\label{app:longlive2}

All measurements use LongLive~2.0's official inference script on an RTX~5090 at $704\times1280$ with identical settings across paths (text encoder on CPU, guidance 1, caches freed before streamed decoding).

\paragraph{Attention and denoising times.} On a 64-s rollout (384 latent frames), the attention path alone takes 78.7\,s with bf16 attention, 84.8\,s with the \kvq{} (78.9\,s bf16 attention, 3.5\,s dequantization, 2.4\,s quantization), 33.5\,s with \sagetwo{}, and 27.3\,s with \ours{}. Whole-network denoising times (medians of five runs) are listed in \cref{tab:speed}: with bf16 weights, the \kvq{} matches plain bf16 (164.8 vs.\ 164.6\,s), while \sagetwo{} takes 119.8\,s and \ours{} takes 107.3\,s. With W4A4 linear layers, comparing all four paths yields 127.2\,s (4-bit release, 10.9\,GiB), 127.0\,s (bf16 attention and cache, 17.9\,GiB), 82.6\,s (\sagetwo{}, 18.1\,GiB), and 70.2\,s (\ours{}, 11.0\,GiB), alongside 126.4 vs.\ 68.8\,s ($1.84\times$) in the primary paired W4A4 run (\cref{tab:speed}).

\paragraph{Memory.} With bf16 weights, peak denoising memory is 24.3\,GiB with the bf16 cache, 17.4\,GiB with the \kvq{}, and 17.5\,GiB with \ours{}. With W4A4 weights, peak denoising memory is 10.9\,GiB for the 4-bit release and 11.0\,GiB for \ours{}, against 17.9\,GiB for bf16 attention and 18.1\,GiB for \sagetwo{}.

\paragraph{VBench-Long.} Under LongLive~2.0's protocol (32 MovieGen prompts, 60-s videos, bf16 weights; \cref{tab:vblong}), neither \ours{} nor the \kvq{} is flagged against bf16 on any of the six dimensions, whereas \sagethree{} (8 prompts) drops in subject consistency, background consistency, and imaging quality. In prompt-level 95\% bootstrap intervals, every interval of \ours{} minus bf16 includes zero, while the \kvq{} lies $0.24$ [$0.06$, $0.45$] points below bf16 in subject consistency; directly paired against the \kvq{}, \ours{} is higher by $0.29$ [$0.12$, $0.48$] in subject consistency, $0.14$ [$0.02$, $0.28$] in background consistency, and $0.07$ [$0.01$, $0.12$] in motion smoothness.

\section{Occasional drops and the robust layout}
\label{app:drops}

\paragraph{Long-rollout stability.} We define a \emph{detail drop} as latent detail in chunks 10--14 (after the window begins sliding) falling more than 3 points below the bf16 reference. On 128 one-minute Self-Forcing rollouts (MovieGen prompts), detail drops occur in 14 rollouts of \ours{} ($10.9\%$), 11 of \sagetwo{} ($8.6\%$), and 11 of a second bf16 run ($8.6\%$); paired McNemar tests show no significant difference between \ours{} and either baseline ($p=0.65$ and $0.66$). When a detail drop is decoded, the picture turns hazy within a few frames at the start of a chunk and stays so.

\paragraph{Chunk-level 8-bit tier.} Extending the 8-bit tier from the newest frame to the full newest chunk is supported as a kernel option at $3.6\%$ additional per-call time and $12.6\%$ larger cache size, useful for deployments that prefer extra margin over maximum throughput. Because the detail-drop rate under standard bf16 initialization is indistinguishable from bf16 and \sagetwo{}, we keep the single-frame 8-bit tier as the default.

\section{Standard metrics, full tables}
\label{app:vbench}

\begin{table}[h]
\caption{\textbf{VBench scores on Self-Forcing} (whole video, score $\times100$, higher is better). Flagged: cells of 21 (7 dimensions $\times$ whole, first, and last third) flagged against bf16. $^\ast$\sagethree{} is evaluated on 128 prompts at 11\,s and 32 prompts at 60\,s.}
\label{tab:vbenchfull}
\centering
\small
\setlength{\tabcolsep}{5pt}
\begin{tabular}{@{}lcccc@{}}
\toprule
 & bf16 & \sagetwo{} & \sagethree{}$^\ast$ & \textbf{\ours{}} \\
\midrule
\multicolumn{5}{@{}l}{\emph{11\,s, 946 VBench prompts}} \\
Subject consistency & 96.18 & 96.19 & 91.99 & 96.15 \\
Background consistency & 96.93 & 96.93 & 94.36 & 96.90 \\
Temporal flickering & 98.01 & 98.01 & 97.75 & 98.02 \\
Motion smoothness & 98.93 & 98.93 & 98.69 & 98.93 \\
Dynamic degree & 35.10 & 35.62 & 38.28 & 34.88 \\
Aesthetic quality & 65.29 & 65.28 & 59.66 & 65.27 \\
Imaging quality & 71.33 & 71.35 & 66.50 & 71.36 \\
Flagged & --- & 0/21 & 10/21 & 0/21 \\
\midrule
\multicolumn{5}{@{}l}{\emph{60\,s, 128 MovieGen prompts}} \\
Subject consistency & 80.76 & 80.87 & 75.40 & 80.63 \\
Background consistency & 87.30 & 87.39 & 84.06 & 87.31 \\
Temporal flickering & 97.58 & 97.51 & 97.54 & 97.52 \\
Motion smoothness & 98.35 & 98.27 & 98.42 & 98.27 \\
Dynamic degree & 35.94 & 35.94 & 53.12 & 35.94 \\
Aesthetic quality & 58.64 & 58.60 & 51.37 & 58.79 \\
Imaging quality & 68.24 & 67.82 & 52.35 & 68.19 \\
Flagged & --- & 0/21 & 14/21 & 0/21 \\
\bottomrule
\end{tabular}
\end{table}

\section{Copy head: further experiments}
\label{app:copyhead}

This section adds to \cref{sec:copyhead}.

\paragraph{Origin and fragility.} The copy head is present in the bidirectional Wan~2.1 across multiple timesteps (\cref{fig:copyhead}a), confirming it is inherited rather than introduced by causal distillation. Its 4-bit error causes roughly an order of magnitude more damage than Gaussian noise of the same norm, acting like a mask on the few keys carrying the copied signal. Wan~2.2 5B has no sharp copy head, only softer previous-frame heads (mass $\approx 0.8$, margin 2--3 nats) that are less fragile at 4 bits. On four emulated Self-Forcing scenes, quantizing this single head to 4 bits loses $0.53$ points of first-frame contrast (vs.\ $0.42$ for the entire operator without the 8-bit newest frame), while exact scores for this head alone reduce the loss to $0.01$; applying the same treatment to a random head (layer 29, head 7) or the highest-entropy head (layer 0, head 8) has no effect. Adding the 8-bit newest frame brings first-frame contrast to $+0.05$.

\paragraph{Across models.} Across five streaming models, keeping the newest cached frame in 8 bits restores $0.07$--$0.47$ points of first-frame contrast.

\paragraph{Concentrated heads.} \citet{li2025kvtuner} argue that heads with concentrated attention tolerate low-bit keys best. Under their own key-only INT4 quantizer, however, the copy head exhibits the largest relative output error of all heads in Self-Forcing ($0.34$--$0.39$, vs.\ a median of $0.014$ over 504 head-calls), and lower entropy correlates with larger error across heads (Spearman $\rho=-0.62$). Whether a query's top key flips tracks the ratio of score error to score margin ($\rho=0.93$--$0.97$) rather than concentration.

\section{Kernel details}
\label{app:kernel}

\begin{figure}[h]
  \centering
  \safefig{0.6\textwidth}{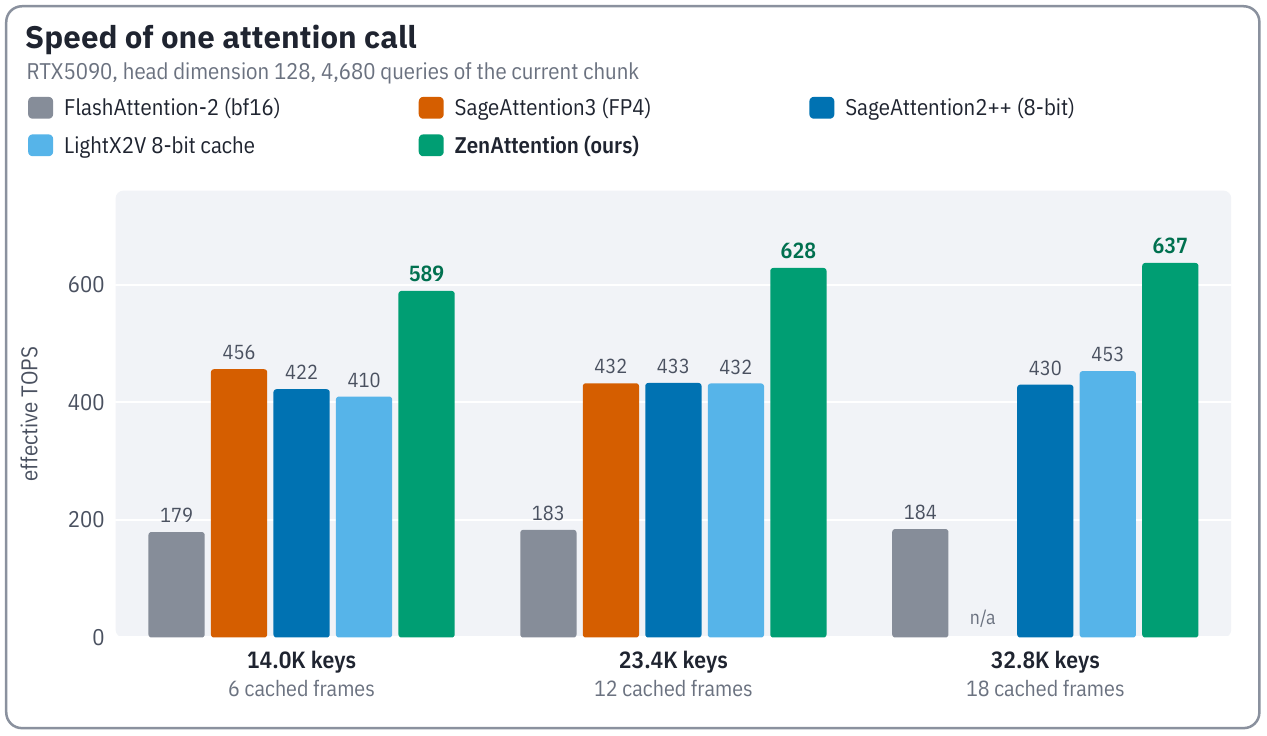}
  \caption{\textbf{Kernel speed} of one call on the Wan~2.1 1.3B streaming shape (4,680 queries; 4,680 current keys plus 1,560 per cached frame) on an RTX~5090; \sagethree{} was not profiled at 18 frames in this sweep.}
  \label{fig:kernel}
\end{figure}

\paragraph{Thread block.} A thread block runs eleven consumer warps, each taking jobs of 16 query rows (an \texttt{m16n8k64} product covers 16 rows), and one producer warp that streams 64-key stages of the records into shared memory with bulk copies. Full and empty \texttt{mbarrier}s synchronize producer and consumers; the stage loop has no block-wide barrier. The kernel supports up to 7{,}168 queries per call and 8 cached chunks.

\paragraph{Instruction counts.} Per consumer warp and FP4 stage (from SASS instruction analysis): 33 tensor-core instructions (535 issue slots); 74 FFMA, 47 LDS, 44 FADD, 42 MUFU, 42 FMNMX, 21 NOP, 20 F2FP, 16 LDSM, and about 112 others; plus about 183 \texttt{mbarrier} wait instructions in the least-loaded warps.

\paragraph{Microbenchmarks.} Pure tensor-core loops in the kernel's tile shape (64 queries, 64-key stages, $d=128$, median of 30 runs, operands in registers) take $1.076\,\mu$s per stage with FP4 operands. INT8 $QK^{\top}$ takes $1.067\,\mu$s and the FP16-accumulated E4M3 $\widetilde PV$ used by \sagetwo{} takes another $1.067\,\mu$s, totaling $2.134\,\mu$s ($1.98\times$); with FP32 accumulation of E4M3 the stage takes $2.771\,\mu$s ($2.58\times$). Loading operands from shared memory yields a $2.45\times$ FP4 advantage, and adding the record stream yields $1.94\times$ (both against FP32 accumulation). Alternative schedules ($128\times128$ tiles, reusing the 8-bit-tier row maximum, and splitting a tile's stages across two thread blocks) stayed within $0\pm2\%$.

\paragraph{Verification and timing.} Kernel times are medians over repeated calls at locked clocks, and per-video self-attention timing synchronizes around every call. The kernel passes \texttt{compute-sanitizer} memory and race checks.

\section{More qualitative comparisons}
\label{app:qual}

\begin{figure}[h]
  \centering
  \safefig{\textwidth}{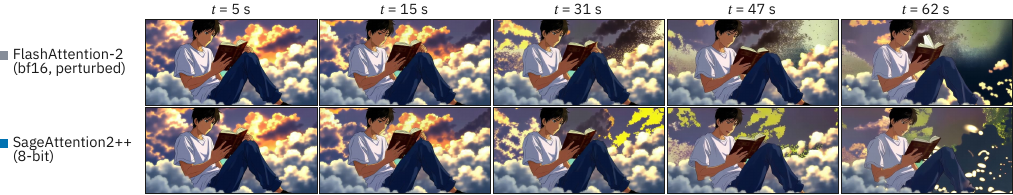}
  \caption{Additional paths for the rollout in \cref{fig:qualtime}: perturbed bf16 (\cref{app:protocol}) and 8-bit attention (\sagetwo{}). The speckled clouds from about 30\,s appear in both.}
  \label{fig:qualread}
\end{figure}

\begin{figure}[h]
  \centering
  \safefig{\textwidth}{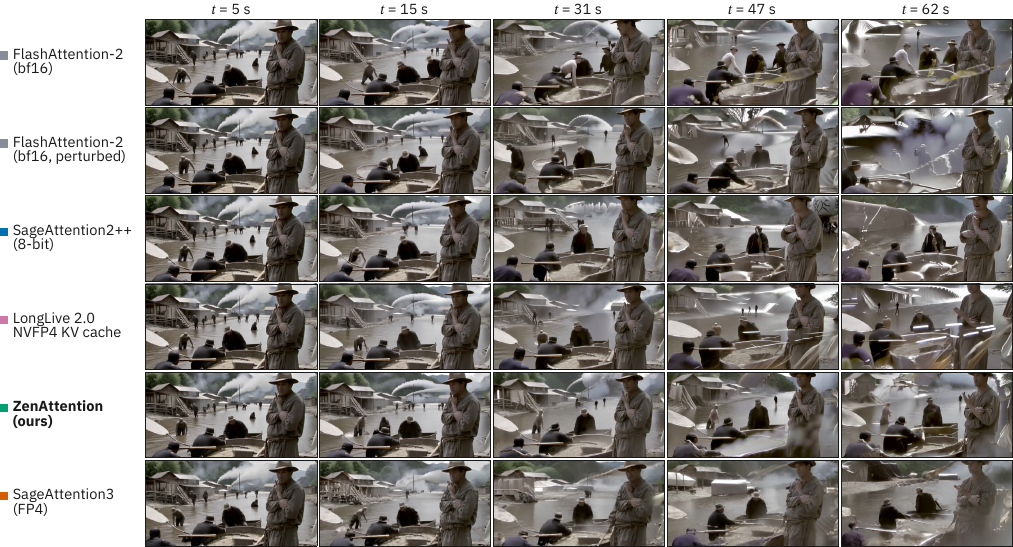}
  \caption{A second LongLive-2.0-5B scene, ``gold-rush town'' (same layout as \cref{fig:qualtime}, all six paths). From about 45\,s all paths exhibit streak artifacts, including both bf16 runs, while \sagethree{} turns hazy starting around 30\,s.}
  \label{fig:qualgold}
\end{figure}

\begin{figure}[h]
  \centering
  \safefig{\textwidth}{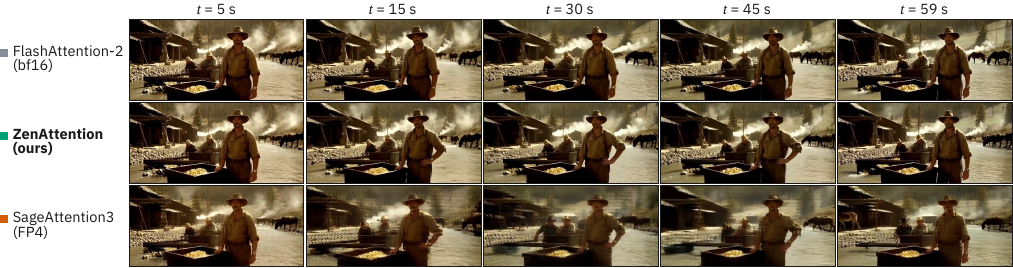}
  \caption{A one-minute rollout of LongLive-1.3B ($480\times832$, ``gold-rush town''; one rollout per row with the same prompt and seed). \sagethree{} turns hazy from about 10\,s on and does not recover; \ours{} follows bf16. Content diverges between rows over time, as it does between two bf16 runs.}
  \label{fig:qualll1}
\end{figure}

\begin{figure}[h]
  \centering
  \safefig{\textwidth}{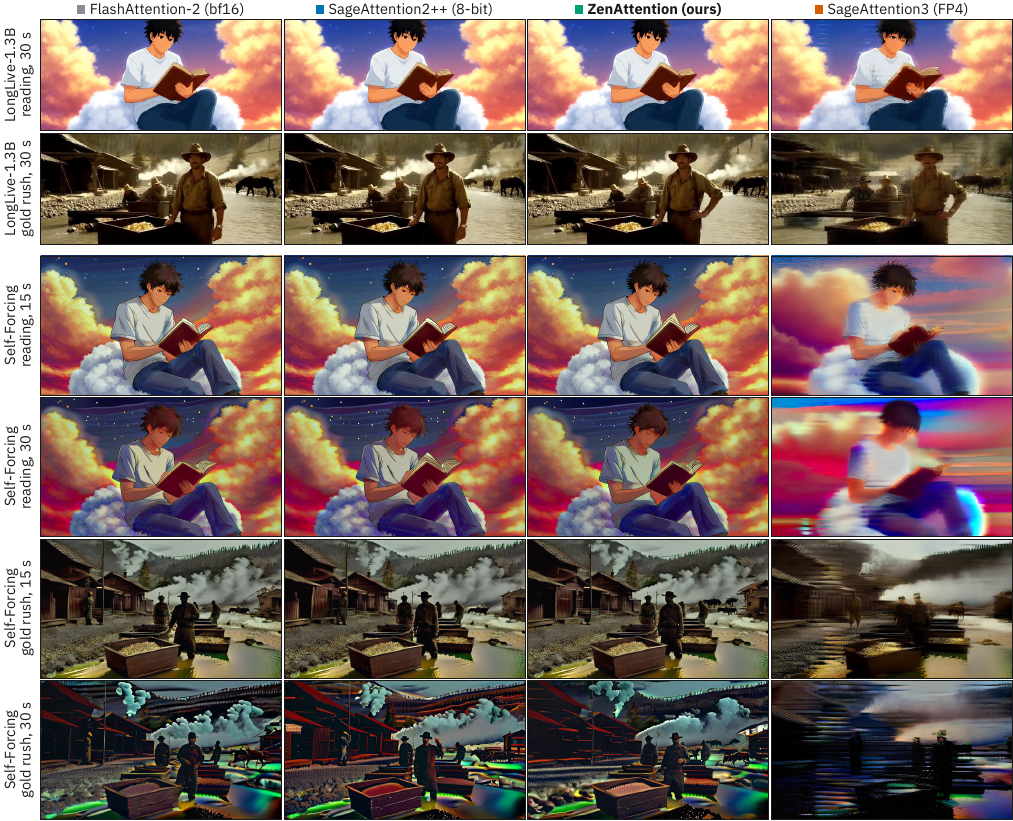}
  \caption{Single rollouts with the same prompt and seed per row: LongLive-1.3B at 30\,s and Self-Forcing (Wan~2.1 1.3B, $480\times832$) at 15\,s and 30\,s, for ``reading on a cloud'' (seed 23) and ``gold-rush town'' (seed 29). On Self-Forcing the bf16 baseline itself oversaturates by 30\,s, beyond the 5\,s it was trained on, and \ours{} follows it; \sagethree{} already collapses into blur and horizontal streaks at 15\,s.}
  \label{fig:qualmore}
\end{figure}

\end{document}